\documentclass[times, twoside]{arxiv}
\usepackage{graphicx}
\graphicspath{ {./img/} }
\usepackage{xspace}
\usepackage{tabularx}
\usepackage{booktabs}
\usepackage{enumitem}
\setlist{nosep}

\usepackage{multirow}
\usepackage{array}
\usepackage{xr}

\definecolor{fgreen}{RGB}{34, 139, 34}

\newcolumntype{P}[1]{>{\raggedright\arraybackslash}p{#1}}

\leadauthor{Li}

\begin{document}

\title{An Extreme-Adaptive Time Series Prediction Model Based on Probability-Enhanced LSTM Neural Networks}
\shorttitle{Extreme-Adaptive Time Series Prediction}

\author[1]{Yanhong Li}
\author[2]{Jack Xu}
\author[1]{David C. Anastasiu\thanks{Corresponding author: David C. Anastasiu. E-mail: \url{danastasiu@scu.edu}}}
\affil[1]{Department of Computer Science and Engineering, Santa Clara University, Santa Clara, CA, USA}
\affil[2]{Santa Clara Valley Water District, San Jose, CA, USA}

\maketitle
%%%%%%%%%%%%%%%%%%%%%%%%%%%%%%%%%%%%%%%%%%%%%%%%%%%%%%%%%%%%%%%%%%%%%%%%%%%%%%%%
\begin{abstract}
Forecasting time series with \textit{extreme} events has been a challenging and prevalent research topic, especially when the time series data are affected by complicated uncertain factors, such as is the case in hydrologic prediction. Diverse traditional and deep learning models have been applied to discover the nonlinear relationships and recognize the complex patterns in these types of data. However, existing methods usually ignore the negative influence of imbalanced data, or severe events, on model training. Moreover, methods are usually evaluated on a small number of generally well-behaved time series, which does not show their ability to generalize. To tackle these issues, we propose a novel probability-enhanced neural network model, called NEC+, which concurrently learns \textit{extreme} and \textit{normal} prediction functions and a way to choose among them via selective back propagation. We evaluate the proposed model on the difficult 3-day ahead hourly water level prediction task applied to 9 reservoirs in California. Experimental results demonstrate that the proposed model significantly outperforms state-of-the-art baselines and exhibits superior generalization ability on data with diverse distributions.
\end{abstract}

% \keywords{multivariate time series analysis, hydrologic prediction, reservoir water level prediction, extreme value theory, deep learning models}

%%%%%%%%%%%%%%%%%%%%%%%%%%%%%%%%%%%%%%%%%%%%%%%%%%%%%%%%%%%%%%%%%%%%%%%%%%%%%%%%
\section*{Introduction}

Time series forecasting is an important technique for many domains in which most types of data are stored as time sequences, including traffic~\cite{hua-aicity2018}, weather forecasting~\cite{hewage2021deep}, biology~\cite{bose-fim2022}, stock price forecasting~\cite{mohan-bds2019}, and water resource management~\cite{Zhang_2021}. These data usually contain seasonality, long term trends, and non-stationary characteristics which usually are taken into account by traditional models during prediction. However, in hydrologic prediction, the water level of dams and reservoirs are also affected by complicated uncertain factors like weather, geography, and human activities, which makes the task of precisely predicting them challenging. Most reservoirs are large hydraulic constructions that serve multiple purposes, including power generation, flood control, irrigation, and navigation, making them critical components in the safety and quality of life of the general population. Therefore, a large number of studies and architectures have explored the problem of reservoir water level prediction.

For a long time, water level prediction was mainly based on traditional machine learning and statistics-based models. However, methods such as Autoregressive Integrated Moving Average (ARIMA)~\cite{boxjen76} seem to adjust poorly to extreme changes in the water level values and cannot easily find the nonlinear relationships among the data. Recently, deep neural networks (DNNs) have shown their great advantages in various areas~\cite{yang2019real,anastasiu-jbdat2020,Yifei2021}. Both conventional Neural Network (NN) and Recurrent Neural Network (RNN) models have been used to overcome the disadvantages of traditional methods for time series forecasting~\cite{Qi2019ADL}, since they can map time series data into latent representations by capturing the non-linear relationships of data in sequences. In particular, Long Short-Term Memory (LSTM) models generally outperform other models in long-term predictions. However, imbalanced data or severe events might hurt deep learning models when it comes to long-term predictions. In the context of reservoir water level forecasting, most of the works mentioned above falter when predicting \textit{extreme} events. Furthermore, they usually focus on predicting only one or two sensors, putting in question the generalizability of these models. To solve these two challenges, we provide an extreme-adaptive solution for reservoir water level prediction which we evaluate extensively on data from 9 different reservoirs across more than 30 years.

% \begin{figure}[b]
%  \centering
%  \includegraphics[width=0.31\textwidth]{reservoir_2.jpg}
%   \caption{Guadalupe Reservoir, CA.}\label{fig:reservoir}
% \end{figure}

The fundamental contribution of this research is the proposal of NEC+, a probability-enhanced neural network framework. We use an unsupervised clustering approach in NEC+ to dynamically produce distribution indicators, which improves the model's robustness to the occurrence of severe events. To improve training performance, we present a selected backpropagation approach and a two-level sampling algorithm to accommodate imbalanced extreme data, along with a customizable weighted loss function for implementing a binary classifier, which is a crucial component of NEC+.

% To sum up, our main contributions are as follows,
% \begin{itemize}
%     \item \textbf{General framework}: We propose an extreme-adaptive time series prediction framework, NEC, based on extreme value theory, which provides better predictions and a flexible combination mechanism to adapt to various occurrences of \textit{extreme} events in reservoir water level predictions. NEC uses historical observations to train, in parallel, a \textbf{N}ormal, an \textbf{E}xtreme, and a \textbf{C}lassification model, relying on dedicated sampling policies and a selected back propagation mechanism to address the data imbalance problem.
%     \item \textbf{NEC+}: Using the NEC framework, we build a probability-enhanced LSTM-based neural network model, NEC+, that uses Gaussian mixture model clustering and optional exogenous variables to enhance the ability of the model to detect and predict \textit{extreme} events.
%     \item \textbf{Effectiveness}: We conduct extensive experiments on real data, which show that NEC+ can significantly outperform state-of-the-art methods by as much as 56\%. NEC+ is able to capture \textit{extreme} events and shows good generalizability on 9 reservoir time series with different distributions over a period of over 30 years. 
% \end{itemize}

\section*{Related work}\label{sec:related}

Time series prediction has been studied extensively. Traditionally, there were several techniques used to effectively forecast future values in the time series, including univariate Autoregressive (AR), Moving Average (MA), Simple Exponential Smoothing (SES), Extreme Learning Machine (ELM)~\cite{6035797}, and more notably ARIMA~\cite{boxjen76} and its many variations. In particular, the ARIMA model has demonstrated it can outperform even deep learning models in predicting future stock~\cite{ariyo2014stock} and dam reservoir inflow levels~\cite{VALIPOUR2013433}. Prophet~\cite{Taylor2018Profet} is an additive model that fits nonlinear trends with seasonal and holiday impacts at the annual, weekly, and daily levels. A number of other classical machine learning models have been used for the task of water level prediction. Due to lack of space, we details them in the additional related work section in the appendix.

% In the specific task of water level prediction, Castillo-Bot{\'o}n et al.~\cite{CastilloBotn2020AnalysisAP} applied support vector regression (SVR), Gaussian processes, and artificial neural networks (ANNs) to obtain short-term water level predictions in a hydroelectric dam reservoir. Several studies by Nayak et al.~\cite{NAYAK200452}, Larrea et al.~\cite{w13152011}, and Tsao et al.~\cite{en14123461} use neural network models (NN) and adaptive neuro-fuzzy inference systems (ANFIS)~\cite{Chang2006AdaptiveNI} to forecast the water level. Some machine learning algorithms, such as Gaussian process regression (GPR) and quantile regression, can not only predict but also quantify forecast uncertainty. Tree-based models are computationally inexpensive and have the advantage that they do not assume any specific distribution in the predictors~\cite{RFbased}. Classification and regression trees (CARTs) and random forest (RF) have also been used to solve hydrologic prediction problems~~\cite{RFbased}. Nguyen et al.~\cite{Nguyen_2021} proposed a hybrid model for hourly water level prediction that integrates the XGBoost model with two evolutionary algorithms and showed that it outperformed RF and CART in the multistep-ahead prediction of water levels. While neural networks were used in some works for water level prediction, they were usually shallow networks that were not able to recognize complex patterns in the data, so feature engineering and extensive manual tuning based on domain expertise had to be employed to improve their performance~\cite{Li01}. 

With the recent success of deep neural network (DNN) models~\cite{hochreiter1997long,fischer2018deep,gers2000learning}, 
hybrid models incorporating different prediction methodologies were also used for water level prediction. Zhang et al.~\cite{Zhang_2021} designed CNNLSTM, a deep learning hybrid model based on the Convolutional Neural Network (CNN) and LSTM models, to predict downstream water levels. Du and Liang~\cite{9656315} created an ensemble LSTM and Prophet model which was shown to outperform any of the single models used in the ensemble. Le et al.~\cite{le2021attention} added an attention mechanism~\cite{xu2015show,chorowski2015attention} to an encoder-decoder architecture to solve the hydro prediction problem.
Even in more traditional time series prediction tasks, Siami-Namini et al.~\cite{8614252} report that deep learning-based algorithms such as LSTM outperform traditional algorithms such as ARIMA given sufficient input data.
Iba\~{n}ez et al.~\cite{w14010034} examined two versions of the LSTM based DNN model exactly for the reservoir water level forecasting problem, a univariate encoder-decoder model (DNN-U) and a multivariate version (DNN-M). Both models used trigonometric time series encoding.
%, but the DNN-M model added several exogenous variables during training, including rainfall, oceanic Ni\~{n}o index, and irrigation releases, some of which we did not have access to during this study. 
% The DNN-U variant showed superior performance in generating short term forecasts compared to other state-of-the-art approaches, and we use it as another baseline in our work. On the other hand, the 
%DNN-M performance only improved the result performance marginally, between 1.5\%--4.5\% in 30 and 90 future event predictions, respectively. 

Statistical methods also provide promising solutions when they are combined with DNN models, especially in the field of sales forecasting. DeepAR~\cite{salinas2020deepar} approximates the conditional distribution using a neural network. Deep State Space Models (DeepState)~\cite{rangapuram2018deep} is a probabilistic forecasting model that fuses state space models and deep neural networks. By choosing the appropriate probability distribution, the bias in the objective function becomes further reduced and the prediction accuracy can be improved. Tyralis and Papacharalampous showed that the architecture can be simply calibrated using the quantile~\cite{tyralis2021quantile} or the expectile~\cite{waltrup2015expectile} loss functions for delivering quantile or expectile hydrologic predictions and forecasts. N-BEATS~\cite{oreshkin2019n}, builds a pure deep learning solution which outperforms well-established statistical approaches in more general time series problems. The N-BEATS interpretable architecture is composed of 2 stacks, namely a trend model and a seasonality model.

While many recent water level prediction methods showed they can outperform traditional or simple DNN models, none of them consider the imbalance of \textit{extreme} vs. \textit{normal} events in the time series and hence ignore the negative influence of \textit{extreme} values on model training. 
Generally, these \textit{extreme} values could be deemed as outliers and be recognized and even removed during data preprocessing. However, 
in our problem, accurate prediction of \textit{extreme} events is generally even more important than the prediction of normal ones. However, we focus on achieving the \textit{best overall prediction performance}, without sacrificing either the quality of \textit{normal} or of \textit{extreme} predictions.
% In contrast, our work contributes to build a delicate design framework NEC+ to solve this problem based on the recognition that \textit{normal} and \textit{extreme} training is hard to obtain the best loss in a single model concurrently. The experiment results show that NEC+ we proposed can surpass the state of the art and shows elegant generalization character. 

\section*{Preliminaries}
\subsection{Problem Statement}
We take on a challenging univariate time series forecasting problem, considering that the data contain a majority of normal values that significantly contribute to the overall prediction performance, along with a minority of extreme values that must be precisely forecasted to avoid disastrous events.

The problem can be described as, 
$$
      [x_1,x_2,\ldots,x_T] \in \mathbb{R}^T \rightarrow[x_{T+1},\ldots,x_{T+H}], \in \mathbb{R}^H
$$
which means predicting the vector of length-$H$ horizon future values, given a length-$T$ observed series history, where $x_1$ to $x_T$ are inputs and $x_{T+1}$ to $x_{T+H}$  are the outputs. %We denote $\widehat{x}$ the forecast of $x$. 
Root mean square error (RMSE) and mean absolute percentage error (MAPE), as standard scale-free metrics, are used to evaluate forecasting performance.

For our experiments, we obtained approximately 31 years of hourly reservoir water level sensor data, along with rain gauge data from a number of sensors in the same area. The Santa Clara reservoirs were built for water conservation in the 1930s and 1950s in order to catch storm runoff that would otherwise drain into the San Francisco Bay. The reservoirs also provide flood protection by controlling runoff early in the rainy season,  recreational opportunities, and they aid the ecology by storing water to keep rivers flowing.

Our models predict 72 (hours) future reservoir water level values, i.e., 3 days ahead. Table~\ref{tbl:data} in the appendix shows the location and type of sensors used in this study. In the remainder of the paper, we will refer to the sensors and theirs associated time series by their given sensor ID in the table.  

The forecast of reservoir water levels is critical for the management of a water resources system. %It can be done hours, days, months, or even years in advance.
Local and state governments need accurate forecasts in order to allocate water resources for competing uses, such as agriculture, domestic household usage, hydropower generation, and environmental objectives. Accurate prediction of extreme events is also essential to prevent reservoir overfill that may have catastrophic flooding results. Reservoir level prediction is especially essential in California, which has had to deal with severe drought for many years in recent decades. 

\begin{figure*}[htb]
 \centering
 \scriptsize
  \includegraphics[width=\textwidth]{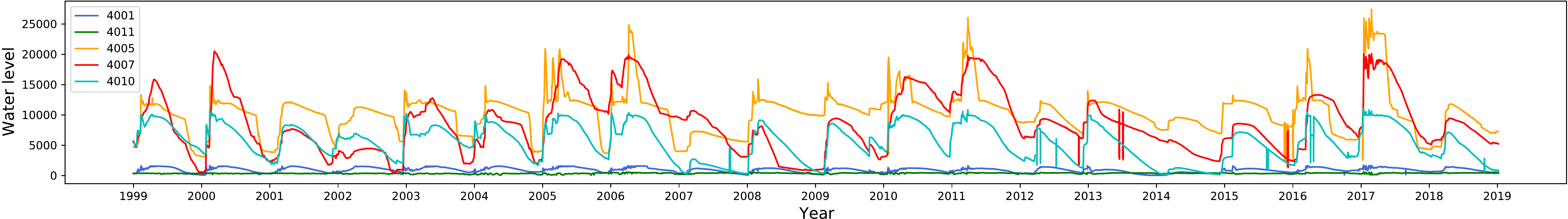}\\
  \caption{Water level values for 5 reservoirs across 20 years.}\label{fig:wave}
\end{figure*}
% Generally, ML model assumes that time-series data consists of some systematic pattern and some random noise. If the random noise is removed, then the systematic pattern would be more prominent. The task of these models is to find the pattern by fitting with training set. However, as we have discussed in related work, these “random noise” can actually be meaningful extreme events like the drought or flood in water system. So, to be more general, we don’t make any assumption of time series data distribution in this paper. Actually, from figure 2, our experiments include varies of reservoir sensors data with different distributions and important extreme events.

\subsection{Extreme Events}

Extreme Value Theory (EVT) tries to explain the stochastic behavior of extreme events found in the tails of probability distributions, which often follow a very different distribution than ``normal'' values. Towards that end, the Generalized Extreme Value (GEV) distribution is a continuous probability distribution that generalizes extreme values that follow the Gumbel (Type I), Fr\'{e}chet (Type II), or Weibull (Type III) distributions. Its cumulative distribution function (CDF) is described as
\begin{equation}
    F(x;\mu, \sigma,\xi) = \exp\left\{-\left[1+\xi\left(\frac{x-\mu}{\sigma}\right)\right]^{-1/\xi}\right\},
\end{equation}
where $\mu\in \mathbb{R}$, $\sigma > 0$, and $\xi$ are the location, scale, and shape parameters, respectively, conditioned on $1+\xi(x-\mu)/\sigma > 0$.
%$\mu\in \mathbb{R}$ is a location parameter, $\sigma > 0$ is a scale parameter, and $\xi$ is a shape parameter, conditioned on $1+\xi(x-\mu)/\sigma > 0$.

Figure~\ref{fig:wave} shows water levels for five of our 9 sensors across a period of 20 years. In order to understand whether extreme events were present in these data, we fit GEV and Gaussian probability density functions (pdf) to the water level values and found that the GEV distribution provides a better fit. In particular, the RMSE of the Gaussian distribution fit is 26.9\%, 46.0\%, and 37.2\% higher than that of the GEV distribution fit for the 4001, 4003, and 4009 reservoirs, respectively, which we also show graphically in the appendix. However, our data has distinct \textit{seasonality} (rain in winter will increase water levels) and \textit{trends} (reservoirs slowly deplete over the year). A time series with trends, or with seasonality, is not stationary and will generally lead to inferior predictions. Therefore, we follow a standard time series analysis preprocessing approach and obtain a stationary time series by applying first-order differencing and then standardizing the resulting time series values, $x_t' = x_t-x_{t-1}$, and $x' = \frac{x'-\mu}{\sigma}$,
% \begin{align*}
% x_t' &= x_t-x_{t-1},\\
% x' &= \frac{x'-\mu}{\sigma},
% \end{align*}
where $\mu$ and $\sigma$ here are the mean (location parameter) and standard deviation (scale parameter) of the Gaussian distribution of the time series $x'$. After obtaining predictions for a time series, we use the same location and scale parameters to inverse the standardization, and the last ground truth value in the time series to inverse the first-order differencing, obtaining values in the same range as the original time series. %The bottom charts in Figure~\ref{fig:wave} show the pdf of the differenced and standardized time series values for the same sensors as in the top and middle figures, along with the best fit Gaussian for those same values. The y-axis is limited to the range $[0,150]$ for visibility but otherwise stretches to $80,000$ for the 0 water level difference bar in the histograms of the three sensors (most of the time there is no change in water level from one hour to the next), resulting in very thin and tall Gaussian distributions.
% The probability density function for the Gaussian distribution with mean $\mu$ and standard deviation $\sigma$ is given by the following formula,
% $$
% \Phi(x;\mu,\sigma) = {\frac{1}{\sigma\sqrt{2\pi}}}\exp{-\frac{{(x-\mu)}^2}{2{\sigma}^2}}.
% $$
When the mean of the distribution is 0 ($\mu=0$) and its standard deviation is 1 ($\sigma=1$), as is the case in our standardized time series, 68\% of the values lie within 1 standard deviation, 95\%  within 2 standard deviations, and 99.7\% within 3 standard deviations from the mean. Yet our time series show values that are up to 100 standard deviations away from the mean in both directions. In our work, we define \textit{normal} values as those within $\epsilon\times\sigma$ of the mean of the preprocessed time series, in both directions, where $\epsilon$ is a meta-parameter we tune for each time series, i.e., $x'_n \in [-\epsilon, \epsilon]$, since $\sigma=1$. The remaining values in the time series are then labeled as \textit{extreme} values.

\section*{Methods}\label{sec:methods}
\begin{figure*}[htb]
 \centering
  \includegraphics[width=0.9\textwidth]{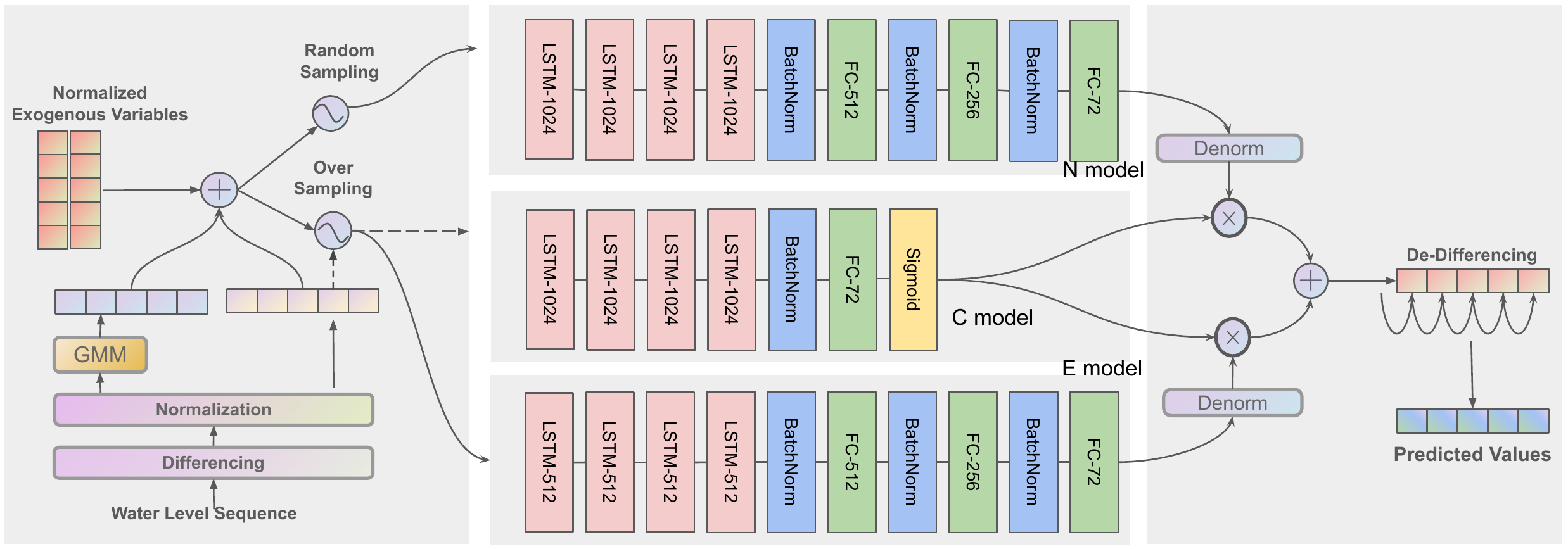}
  \label{NEC+ Architecture }
  \caption{The NEC+ framework; $\bigoplus$ denotes the element-wise addition and $\bigotimes$ denotes the element-wise product.}\label{fig:framework}
\end{figure*}

\subsection{NEC}\label{sec:methods:framework}
We designed our NEC framework to account for the distribution shift between \textit{normal} and \textit{extreme} values in the time series. NEC is composed of three separate models, which can be trained in parallel. The Normal (N) model is trained to best fit \textit{normal} values in the time series, the Extreme (E) model is trained to best fit \textit{extreme} time series values, and a third Classifier (C) model is trained to detect when a certain value may be categorized as \textit{normal} or \textit{extreme}. The framework is flexible and may use any prediction models for the 3 components, yet in this work, given the evidence presented in the related work section, we focus on deep learning models that use a fixed set of $h$ consecutive past values as input to predict the next $f$ values in the time series. At prediction time, the C model is used to decide, for each of the following $f$ time points, whether the value will be \textit{normal} or \textit{extreme}, and the appropriate regression model is then applied to obtain the prediction for those points.

The middle section of Figure~\ref{fig:framework} shows the configuration for our chosen N, E, and C models. The N and E models each have 4--6 LSTM layers followed by 3 fully connected (FC) layers that consecutively reduce the width of the layer down to $f$, which is 72 in our case (3 days). The number of inputs was set to 15 days, i.e., $h=15\times24=360$. Since there are much fewer extreme values in the data than normal ones, we set the LSTM layer width to only 512 in the E model, while we set it to 1024 in the N model. Finally, the C model uses the same size LSTM layers as the N model, followed by a 72 node FC layer with a Sigmoid activation function.

\subsection{GMM Indicator} \label{sec:methods:gmm}

\begin{figure}[htb]
 \centering
  \includegraphics[width=\linewidth]{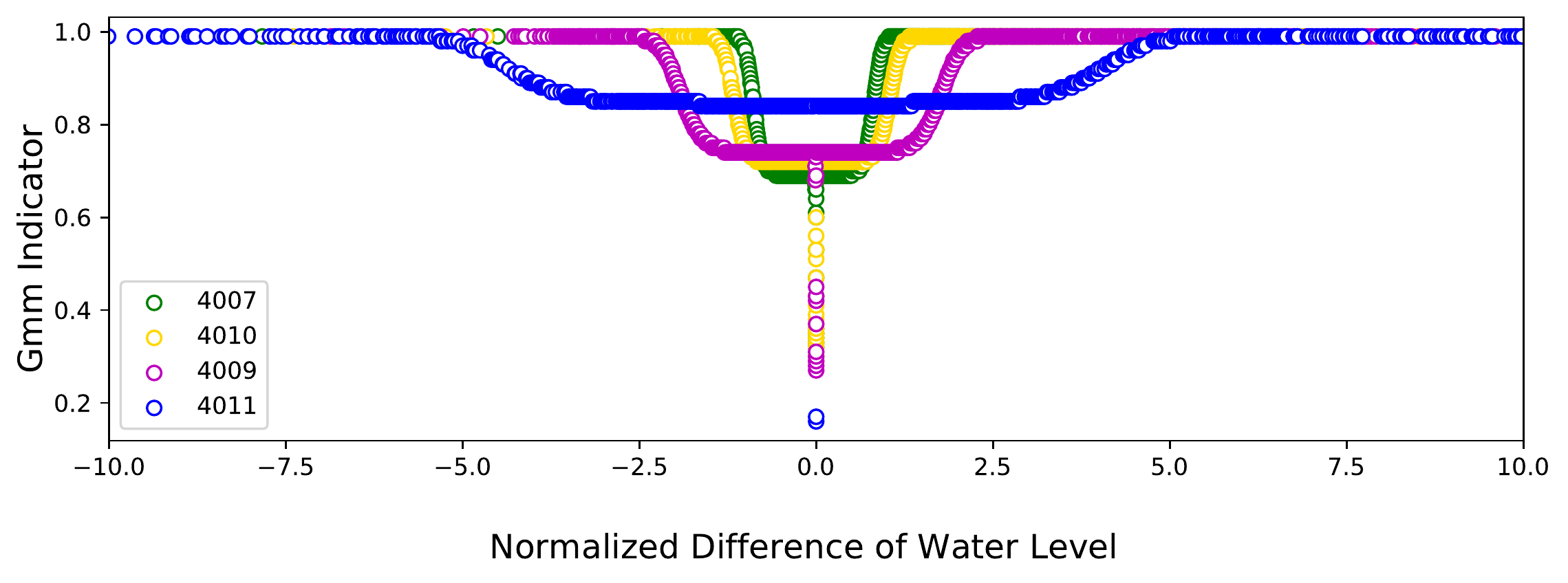}
  \caption{GMM indicator distribution.}\label{fig:gmm}
\end{figure}

A Gaussian mixture model (GMM)~\cite{day1969} can be described by the equation,
$$
p(x|\lambda)=\sum_{i=1}^M w_i~g(x|\mu_i,\mathbf{\Sigma_i}),
$$
where $x$ is a $D$-dimensional continuous-valued vector, $w_i\ \forall i = 1, \ldots, M$ are the mixture weights, and $g(x|\mu_i, \Sigma_i)$, are the component Gaussian densities. Each component density is a $D$-variate Gaussian function, and the overall GMM model is a weighted sum of M component Gaussian densities,
\begin{footnotesize}
$$
g(x|\mathbf{\mu_i},\mathbf{\Sigma_i})=\frac{1}{2\pi^\frac{D}{2}|\Sigma_i|^\frac{1}{2}}\exp{\left\{-\frac{1}{2}{(x-\mu_i)}^T~ \Sigma_i^{-1} (x-\mu_i)\right\}},
$$
\end{footnotesize}
where $\mu_i$ is the mean vector and $\Sigma_i$ is the covariance matrix of the $i$th component. The mixture weights are constrained such that $\sum_{i=1}^M w_i=1$. 
%Due to its capacity to represent a vast class of sample distributions, GMMs are frequently employed in biometric systems, most notably in speaker recognition systems. 
The GMM's capacity to produce smooth approximations to arbitrarily shaped densities is one of its most impressive features~\cite{day1969}. 

In our work, we use Expectation-Maximization to fit a GMM model using the time series training data. Then, each model component can generate a probability for each point in the time series. Finally, for each value in the time series, we compute an \textit{indicator} feature as the weighted sum of all component probabilities, given the weights learned when fitting the GMM model. In our framework, the number of components $M$ is a hyper-parameter which we tune for each time series. As an illustration, \figurename~\ref{fig:gmm} shows the indicator values for GMM with $M=4$ for Sensor 4009 and with $M=3$ for the other 9 sensors. The x-axis in the figures represents the preprocessed time series input, which we limited to the $[-10,10]$ range for visibility, while the y-axis represents the indicator values. It is easy to see that the learned \textit{normal} and \textit{extreme} indicator bounds vary depending on the sensors. We hypothesize that providing this indicator as an additional input feature for our models will help differentiate between \textit{normal} and \textit{extreme} values and minimize prediction errors. Therefore, we extend our N, E, and C models by providing both the water level and its associated GMM indicator as input for each of the $h$ input values.

\subsection{Exogenous Variables} \label{sec:methods:shed}

For some time series, we may provide additional exogenous inputs which may help improve the overall prediction of future values. For example, rain fall in the region around the reservoir is not affected by the reservoir water level but, when it is raining, it can be a strong indicator that the reservoir level may increase soon, as water drains into streams and rivers that may flow into the reservoir. For a given region, a \textit{watershed} is a land area that channels rainfall and snowmelt to creeks, streams, and rivers, and eventually to outflow points such as reservoirs, bays, and the ocean. In our work, as shown in Table~\ref{tbl:data} in the appendix, we define several watersheds and use several rain gauge sensors in those watersheds as exogenous variables to aid in the predictions associated with several of the reservoirs, which were chosen in consultation with domain experts. Namely, for reservoir 4005 we used rain sensor 6017, for 4010 we used 6135, and for 4007 we used both 6044 and 6069. 

\subsection{NEC+} \label{sec:methods:nec-plus}

Our NEC+ framework is described in \figurename~\ref{fig:framework}. Unlike the base NEC framework, which relies only on historical time series values for future predictions, NEC+ adds GMM indicator and watershed exogenous variables (when available) to create multivariate regression N and E models. Given $k$ watershed variables, the NEC+ models will be $k+2$-variate models, after adding the original input and the GMM indicator. In addition, to account for the differences between the distributions of the \textit{normal} and \textit{extreme} values during training, we define custom sampling policies, regression backpropagation, and classification loss function, which we present in the following.
% \begin{algorithm}[htb]
% \caption{The proposed NEC+ method.}
% \textbf{Input:} Reservoir water level series,\\ $k$ watershed series, $\epsilon, M $\\
% \textbf{Output:} Test set forecast 
% \begin{algorithmic}[1]
% \State $D \gets $ Differencing on reservoir water level series
% \State $S, Inp \gets $ Standardization on D
% \If{ Watershed group is not null} 
%   \ForAll { Rain series $\in$ Watershed group}
%     \State $RS \gets $ Standardization on Rain series
%     \State $Inp \gets $ $Inp$ concatenate $RS$
%   \EndFor
% \EndIf
% \State $Val,Te \gets $ Sampling in Validation and Test Years
% \State $Tr \gets $ Random Sampling on $Inp$ except Val and Te
% \State $GMM \gets $ Training GMM with $Inp$ and $M$
% \State $GTr, GVal, GTe \gets $ Inference Tr, Val and Te with GMM
% \State $Tr, Val, Te \gets Tr, Val, Te$ concatenate with $GTr, GVal, GTe$
% \State $NInp, EInp, CInp \gets $ OverSampling on $Tr$ with $\epsilon $
% \State $N,E,C  \gets $ Training $N,E,C$ models with $TrInp, EInp, CInp, Val$
% \State $TestN, TestE, TestC \gets $ Inferencing $Te$ on $N,E,C$ model
% \State $TestN, TestE \gets $ De-Standardized $TestN, TestE$
% \State $TPred \gets  TestN\bigotimes TestC \bigoplus TestN\bigotimes TestC$
% \State \Return $TPred$
% \end{algorithmic}\end{algorithm}

\subsection{Sampling Policies}\label{sec:methods:sampling}
% \begin{table}[htbp]
% \centering
% %\small
% \footnotesize
% \setlength{\tabcolsep}{2mm}{}
% \caption{Over Sampling for Sensor 4005}\label{tbl:over-sampling}
% \begin{tabular}{cccccc}
% \hline
% \textbf{Models} & \textit{RMSE} & OS=0.3 & OS=0.2 & \textbf{OS=0.04} & OS=0 \\
% \hline
% \multirow{3}{*}{L+G} & Total & 22495.9 & 9558.5 & \textbf{7919.8} & 8208.4 \\
% & Normal & 21788.4 & 8894.5 & 6872.9 & 7162.1 \\
% & Extreme & 1992.1 & 1558.0  & 2474.4 & 2478.7 \\
% \hline
% \multirow{3}{*}{L} & Total & 68442.4 & 12300.9 & \textbf{7769.6} & 8183.0 \\
% & Normal & 68170.3 & 11556.7 & 6784.9 & 7122.4 \\
% & Extreme & 917.1 & 1932.6 & 2654.3 & 2512.0 \\
% \hline
% \end{tabular}%
% \end{table}%
Our models require $h$ values from the time series to predict the following $f$ values, and $h,f \ll |x|$, the length of the time series. Moreover, while the number of \textit{extreme} values differs based on the choice of $\epsilon$, it is still quite small in comparison to the number of normal values. In our experiments, $h=360, f=72$, $|x|\sim 276K$, and \textit{extreme} values ranged from 0.08\% to 4.08\% of the time series values across our 9 sensors. Therefore, sampling plays a crucial role during training. However, oversampling cannot be used to mitigate this problem. In an experiment we detail in the appendix (due to lack of space), we found that, while oversampling extreme events improves predictions in that area, it leads to worse overall predictions for the rest of the time series.

% To showcase the importance of sampling, we ran an experiment in which we trained an LSTM model and one that also used GMM features (L and L+G in Table~\ref{tbl:over-sampling}), and allowed different levels of oversampling; OS=0.04 means that 4\% of the training samples had at least one \textit{extreme} value in the prediction area of the sample, even though only 0.5\% of the values in sensor 4005 were deemed \textit{extreme}. Table~\ref{tbl:over-sampling} shows the total RMSE as well as component RMSE scores computed only for the \textit{normal} and \textit{extreme} values. While the extreme values RMSE continues to decrease as the OS level increases, oversampling lead to marginal improved total RMSE scores at OS=0.04 and markedly worse results for higher levels of oversampling. This shows that oversampling cannot be used as a panacea to account for the rarity of the \textit{extreme} values in the time series. Instead, our NEC framework separates the task of predicting \textit{extreme} and \textit{normal} values, achieving markedly improved results in the process.

When training our NEC+ model, we apply a two-stage sampling policy. First, given the high cardinality of our time series, we randomly sample subsections of length $h+f$ from the series as samples to use in training our models, while avoiding sections included in the test and validation sets. Specifically, the validation and test sets each include 24 randomly chosen $f$-length sections from the years 2014 and 2016 for the validation and 2017 and 2018 for the test set, respectively, and the training set includes all other values in the time series. Second, we perform stratified sampling of regions with and without \textit{extreme} values, allowing the E and C models to oversample up to OS\% samples with at least 1 \textit{extreme} value in the prediction zone.

\subsection{Selected Backpropagation in the N and E models}\label{sec:methods:backprop}
\begin{figure}[htb]
  \centering
  \includegraphics[width=\linewidth]{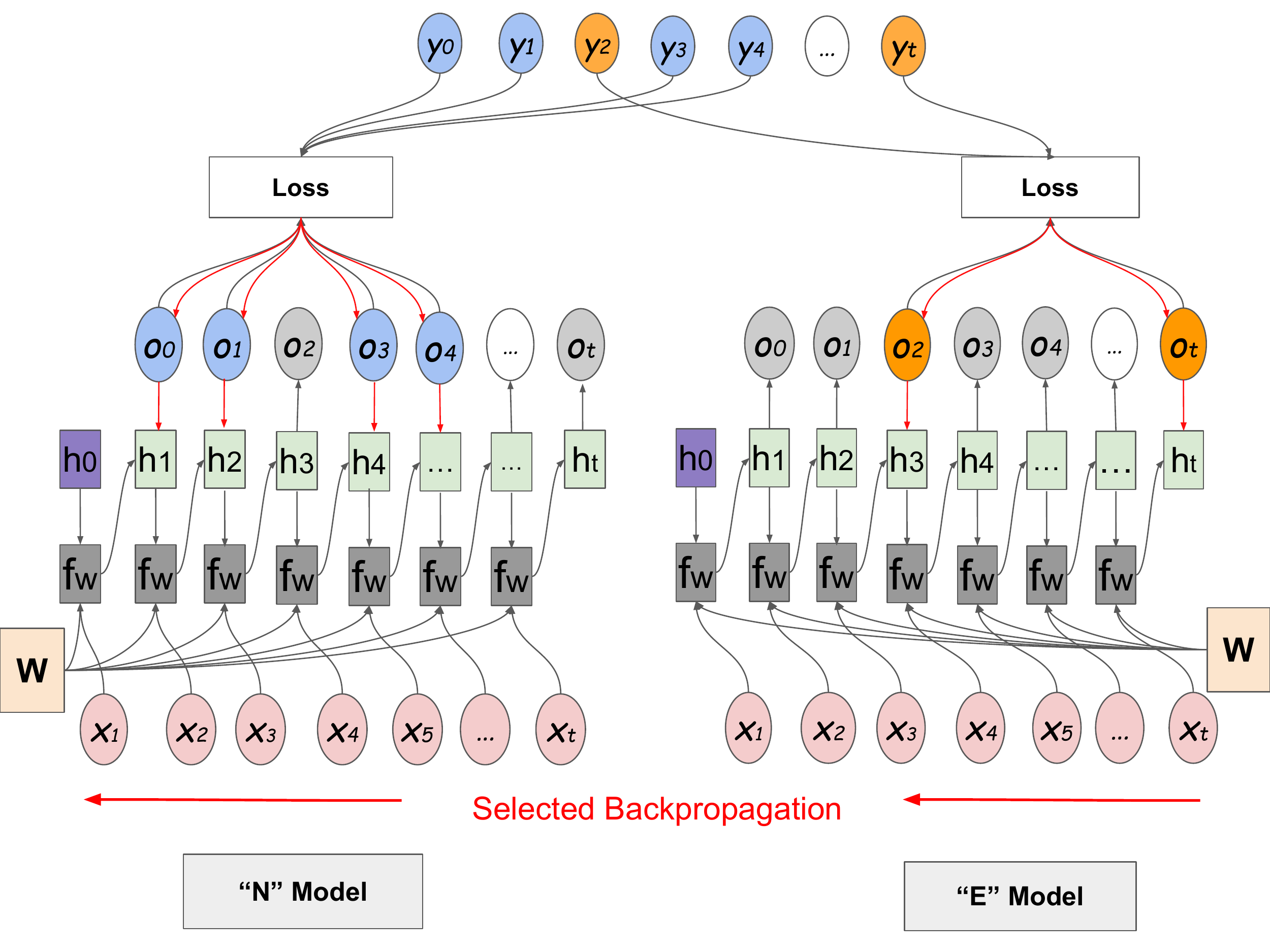}
  \caption{Computational graph for the N and E models. Blue and orange inputs represent \textit{normal} and \textit{extreme} values, respectively.}\label{fig:backprop}
 \end{figure}

In addition to a custom sampling policy, one important approach we use to ensure proper training of the N and E models is \textit{selected backpropagation}, which we describe visually in \figurename~\ref{fig:backprop}. 
Each prediction sample in our data contains $f$ values, only a few of which may be extreme. The rarity of extreme events would cause the E model to be unduly influenced by the loss on normal values, and vice-versa. As a result, our backpropagation ignores predictions on normal values in the E model and on extreme values in the N model, forcing the model to only focus on the values important for the given model.
Specifically, when training the N model, only \textit{normal} values add to the loss, and when training the E model, only \textit{extreme} values add to the loss. This is equivalent to perfect predictions (predicting the ground truth) for \textit{normal} values when training the E model, and perfect predictions for \textit{extreme} values when training the N model. In this way, only the positions and values of appropriate \textit{normal} or \textit{extreme} data will affect the hidden parameters in the network during backpropagation when training the N and E models.

\subsection{Parameterized Loss Function in the C Model}\label{sec:methods:loss}

The Binary Cross Entropy (BCE) loss is usually the most appropriate loss function for binary classification tasks. Based on BCE, we propose a tunable loss function to accommodate the serious imbalance problem in the prediction of time series with extreme events.

BCE loss compares the target, which in our case is whether the value is \textit{normal} (0) or \textit{extreme} (1), with the prediction, which takes values close to 0 or 1 after the transformation of of the Sigmoid function.
The loss increases exponentially when the difference between the prediction and target increases linearly. It can be defined as
\begin{equation}\label{eg:BCE}
BCE(t,p)=-(t\times\log{(p)}+(1-t)\times\log{(1-p))},
\end{equation}
where $t$ and $p$ are the target and predicted values, respectively. However, for datasets with a high imbalance between the two classes, such as our time series, BCE will favor the prominent class. To solve this problem, we propose a parameterized tunable loss as follows,
\begin{equation}
    \mathcal{L} = \beta \times BCE(t,p^\alpha) + (1-\beta) \times {RMSE(t,p)},
\end{equation}
where $\alpha$ and $\beta$ are parameters that can be tuned. Values $\alpha>1$ cause the model to predict $p$ values that are higher in general in order to minimize the distance between $t$ and $p^\alpha$. The BCE part of the loss can be thought of as a blunt instrument that grossly exaggerates all miss-classifications in order to more accurately predict the obscure class, while the RMSE part allows for a more gentle penalty based on the distance between $t$ and $p$. In other words, the higher $\alpha$ is set, the more \textit{extreme} (class 1) predictions can be obtained. The $\beta$ meta-parameter controls the strength of the two components of the loss. For time series that are more balanced, $\beta$ can be small, or even 0. 

\section*{Evaluation}\label{sec:eval}

In this section, we present empirical results for our proposed framework. We are interested in answering the following research questions with regards to prediction effectiveness:
\begin{enumerate}[nosep]
\item What is the effect of adding the GMM indicator to a model?
\item What is the effect of introducing exogenous features?
\item How do the loss function parameters affect performance?
\item How does NEC+ compare against state-of-the-art baselines?
\end{enumerate}
In the following, we will first present the experimental setup and baseline approaches we compared against, and then answer the proposed research questions, in order.
% In particular, the main research questions are: Is our proposed framework effective in time series prediction? How the effect of our method can be improved with multiple Exogenous information? 
% To answer these questions, we organized the experiments into 3 groups:
% \begin{enumerate}
% \item
% To compare the baseline models with NEC+, the 10 reservoir’s own history data, to be specific, the standardized first difference of water level, and self-contained relevant information like date, are used in forecasting each of them respectively. 
% \item
% We further explore the effects of our proposed model on multi dimension inputs, by combining water level with the precipitation’s sensor data within the same watershed on reservoir 4005, 4007 and 4010.  
% \item
% We also examine the extensibility of NEC+ by replacing the predicting unit from stacked LSTM to LSTM-based AutoEncoder on reservoir 4006.
% \end{enumerate}

\subsection{Experimental Settings}

\subsubsection{Dataset}\label{sec:eval:settings:dataset}
Our dataset includes over 31 years of hourly sensor readings for the water level in 9 reservoirs and 5 rain sensor gauges in Santa Clara County, CA, which are described in the appendix and listed in Table~\ref{tbl:data}. After reducing all time series to a common date range, each reservoir and rain sensor time series has 276,226 values. When training all baseline models and the N and C models in NEC+, at least 100,000 random samples were selected, with replacement, from the training set. However, due to the sparsity of \textit{extreme} events, only 50,000 random samples were selected, with replacement, when training the E model. The N model did not use any oversampling ($OS=0$), but we set $OS=1$ for both the E and C models, ensuring that all training samples had at least 1 \textit{extreme} event in the prediction section of the sample. The data for the 9 reservoirs and 4 rain gauge sensors we used in this study, along with the code for this work have been made available on GitHub at \url{https://github.com/davidanastasiu/NECPlus}.

\subsubsection{Model Parameters}\label{sec:eval:settings:params}

For reservoir 4009, we set $M$%, the number of GMM components, 
to $4$ and $\epsilon$%, the extreme value threshold, 
to $1.8$. For all other reservoirs, $M=3$ and $\epsilon=1.5$. For each reservoir, we tested models with 4 or 6 LSTM layers, and 5 reservoirs use 6 LSTM layers while the rest use 4. We also tested LSTM layer widths of 512 and 1024 nodes and found 1024 node layers were better suited for the N and C models, while E models performed better with 512 nodes across all reservoirs. While $f=72$ (3 days) was set by our problem definition, we tested $h \in {72, 168, 360, 720}$, i.e., ${3, 7, 15, 30}$ days, and found $h=360$ to work the best for all reservoirs. 

All models were trained using PyTorch 1.9.1+cu102 on a Linux server running CentOS 7.9.2009 equipped with 2x 20-core Intel(R) Xeon(R) Gold 6148 CPUs, 768 GB RAM, and 3 NVIDIA V100 GPUs.  Finally, the LSTM layers were trained using an SGD optimizer with learning rate 1E-3, while the fully connected layers were trained using an Adam optimizer with learning rate 5E-4.

%\subsubsection{Metrics}\label{sec:eval:settings:metrics}

%In our experiments, we are primarily interested in prediction performance, which we measure via root mean square error (RMSE) and mean absolute percentage error (MAPE) on the samples of the held-out test set, which are then summed up across all test samples. During training, we also ensure proper model training by verifying that the validation set RMSE is not much higher than the training set RMSE given our chosen model meta-parameters.

\subsection{Baseline Methods}\label{sec:eval:methods} 
We compared our proposed method, NEC+, against a wide array of traditional and state-of-the-art time series and reservoir level prediction methods, which are introduced in the related work section and summarized below.
% \begin{itemize}
%     \item ARIMA~\cite{boxjen76}, a standard statistics-based time series analysis method,
%     \item Prophet~\cite{Taylor2018Profet}, a powerful non-linear regression technique which accounts for seasonality,
%     \item LSTM, which is the standard LSTM~\cite{hochreiter1997long} recurrent neural network with the same configuration as our \textit{normal} model,
%     \item DNN-U~\cite{w14010034}, a state-of-the-art univariate LSTM-based encoder-decoder \textit{hydrologic} model used to predict reservoir lagged water levels,
%     \item Attention-LSTM~\cite{le2021attention}, a state-of-the-art \textit{hydrologic} model used to predict stream-flow, and
%     \item N-BEATS~\cite{oreshkin2019n}, a state-of-the-art time series prediction method that outperformed all competitors on the standard M3~\cite{MAKRIDAKIS2000451}, M4~\cite{makridakis2018m4} and TOURISM~\cite{athanasopoulos2011tourism} datasets. %~\rnote{The interpretable architecture is composed of 2 stacks, trend model and seasonality model.}
% \end{itemize}
\begin{itemize}
\item \textbf{ARIMA}~\cite{boxjen76} is sometimes referred to as the Box-Jenkins method. The I in the model refers to integrating the time series values during one or more differencing or seasonal differencing steps, until the series becomes stationary with respect to its means. Then, predictions are performed via an autoregressive moving average model $ARMA(p, q)$ that expresses the future value of a variable as a linear combination of $p$ past values while minimizing $q$ past errors. 

\item \textbf{Prophet}~\cite{Taylor2018Profet} is a time series prediction method based on an additive model that fits nonlinear trends with seasonal and holiday impacts at the annual, weekly, and daily levels. It works well with time series data with strong seasonal influences as well as historical data with many seasons, and it can usually handle outliers. 
% Prophet is a decomposable time series model made up of three components, trend, season, and holidays, which can be modeled as
% $$
%     y(t)=g(t)+s(t)+h(t)+{\epsilon_t}.
% $$
% Here, $g(t)$ is a time series simulation tool that models non-periodic changes, $s(t)$ represents the periodic term, and uses Fourier series to construct a flexible periodic model, and $h(t)$ is a non-cyclical ``holiday'' item whose impact is considered to be predictable to a certain extent. The interference term $\epsilon_t$ represents errors, or changes that the model cannot adapt to, and is generally considered to follow the Gaussian distribution.

\item \textbf{LSTM}~\cite{hochreiter1997long} was first proposed by Hochreiter and Schmidhube. The architecture makes it easier for a neural network to preserve information over many time steps. It does not guarantee that there is no vanishing/exploding gradient, but, compared with traditional recurrent neural networks, it does provide an easier way for the model to learn long-distance dependencies, by using a forget gate, an output gate, and the input gate. We compare our NEC+ model against LSTM and also used 4--6 LSTM layers within our model. When comparing against LSTM, for each sensor, we used the same number of layers for the baseline LSTM model as for our NEC+ model.

\item \textbf{DNN-U}~\cite{w14010034} is a state-of-the-art univariate LSTM-based encoder-decoder \textit{hydrologic} model used to predict reservoir lagged water levels. DNN-U closely follows an earlier sequence-to-sequence (\textit{seq2seq}) architecture proposed by Sutskever et al.~\cite{seq2seq}. The encoder LSTM takes in lagged observations of the water level $y$ and encodes them into a hidden state $h$ and cell state $c$. These encoder states are then used as the initial states for the decoder LSTM, which accepts known future dates $d$ as input, representing the target dates they wish to forecast. The decoder outputs are then passed to a time distributed dense layer, which generates the actual forecasts.

\item \textbf{Attention-LSTM}~\cite{le2021attention} is a \textit{hydrologic} prediction model that forecasts the stream flow values over the next 12 hours, relying on an attention mechanism based on LSTM networks that consider past stream data. The context vector used as the decoder input is determined by the last hidden state of the encoder. %The specialty of the Attention mechanism of the Attention-LSTM model is that different weights are given to the inputs according to the output, and the LSTM network plays a role of dynamically retaining the input weights for a long time.

\item \textbf{N-BEATS}~\cite{oreshkin2019n} is a deep neural architecture which is composed of groups of residual links and a deep stack of fully-connected layers. It is a state-of-the-art network on several well-known datasets, including M3, M4 and TOURISM competition datasets. It was compared with DeepAR~\cite{salinas2020deepar}, and Deep State~\cite{rangapuram2018deep}, the winner of the M4 competition~\cite{makridakis2018m4}, which represents a hybrid approach between neural network and statistical time series models.
\end{itemize}

\subsection{Effect of Adding the GMM Indicator Variable}

\begin{figure}[htb]
\centering
% \captionsetup[subfloat]{farskip=2pt,captionskip=1pt}
\includegraphics[width=0.325\linewidth]{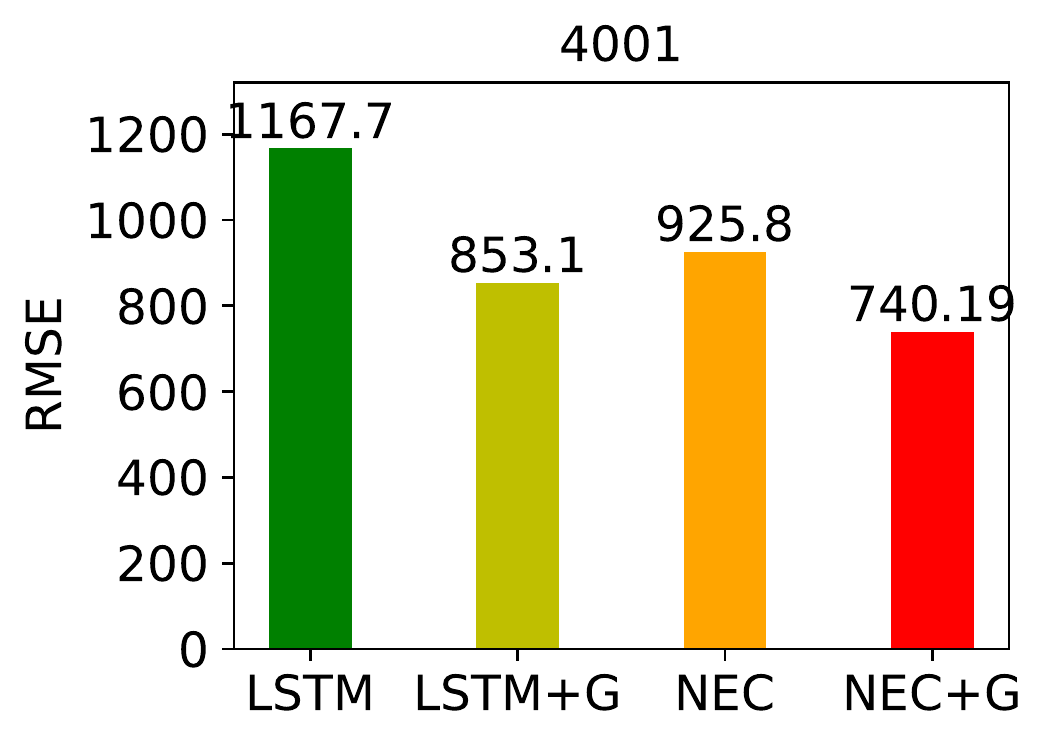}
\includegraphics[width=0.325\linewidth]{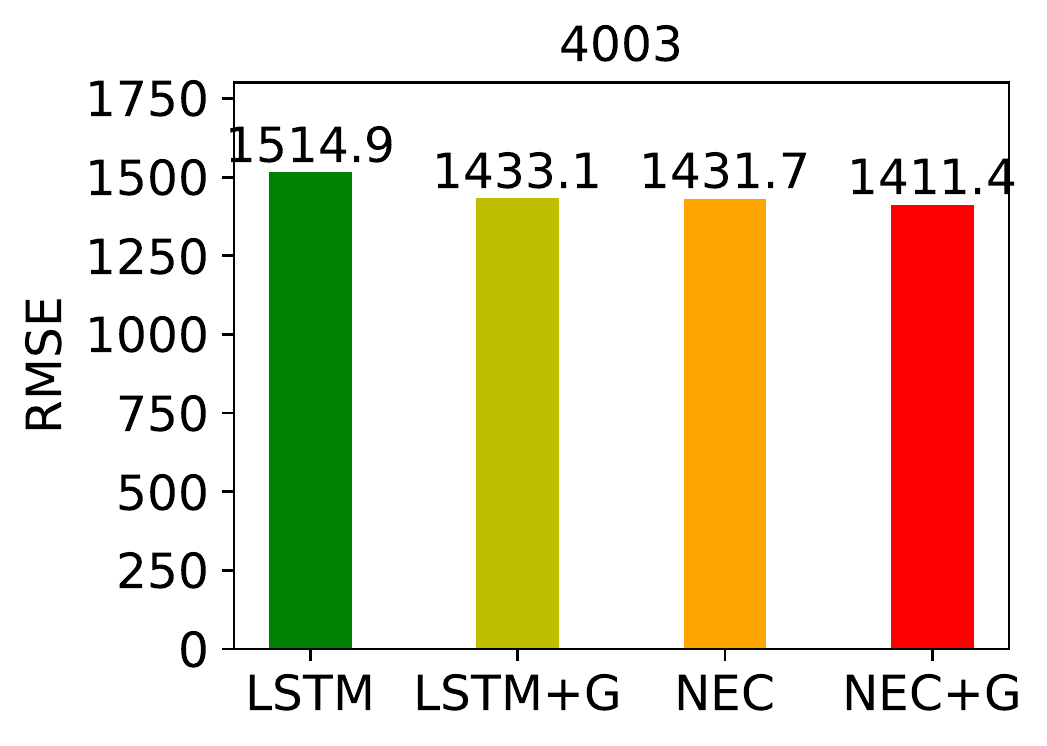}
\includegraphics[width=0.325\linewidth]{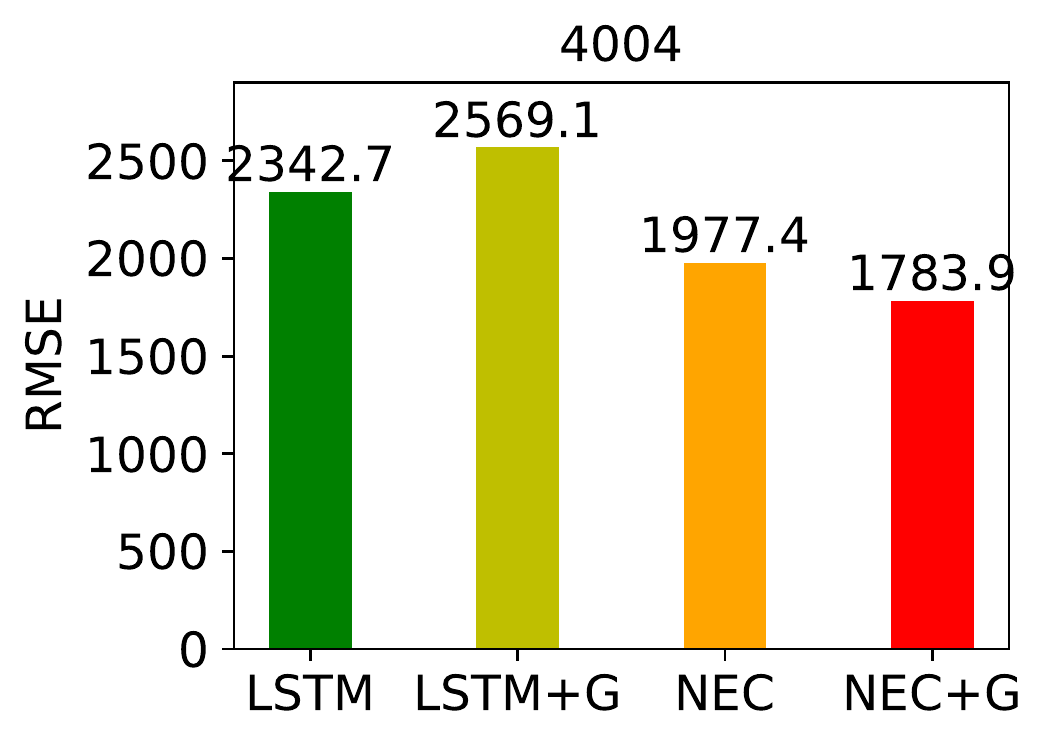}\\
\includegraphics[width=0.325\linewidth]{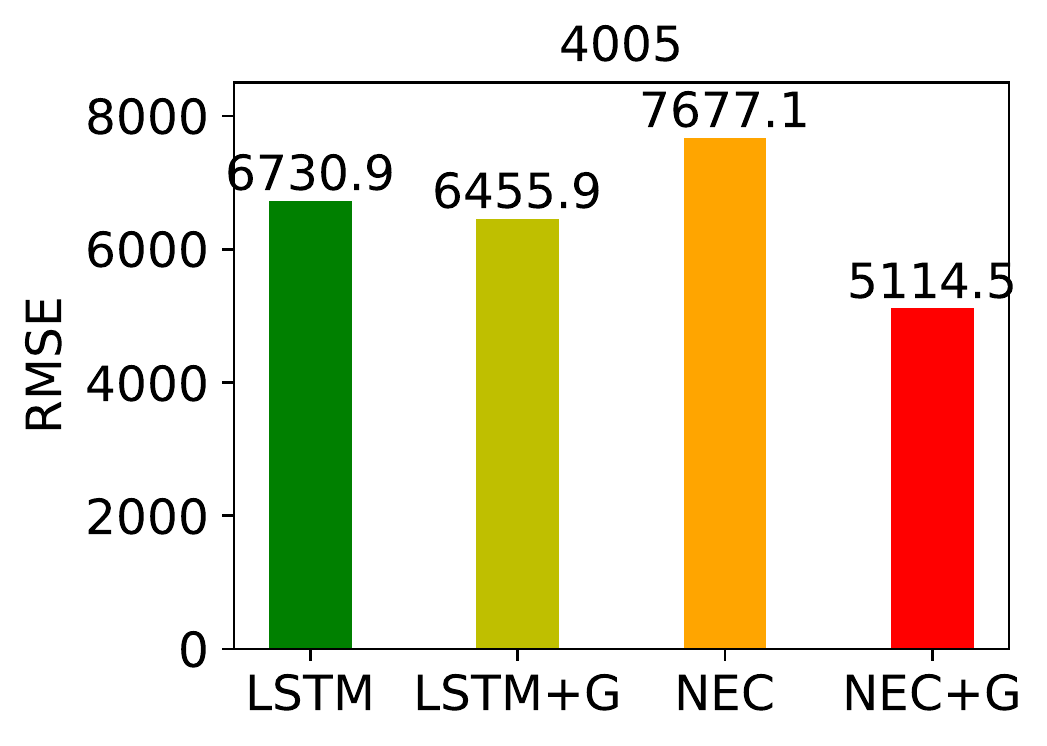}
\includegraphics[width=0.325\linewidth]{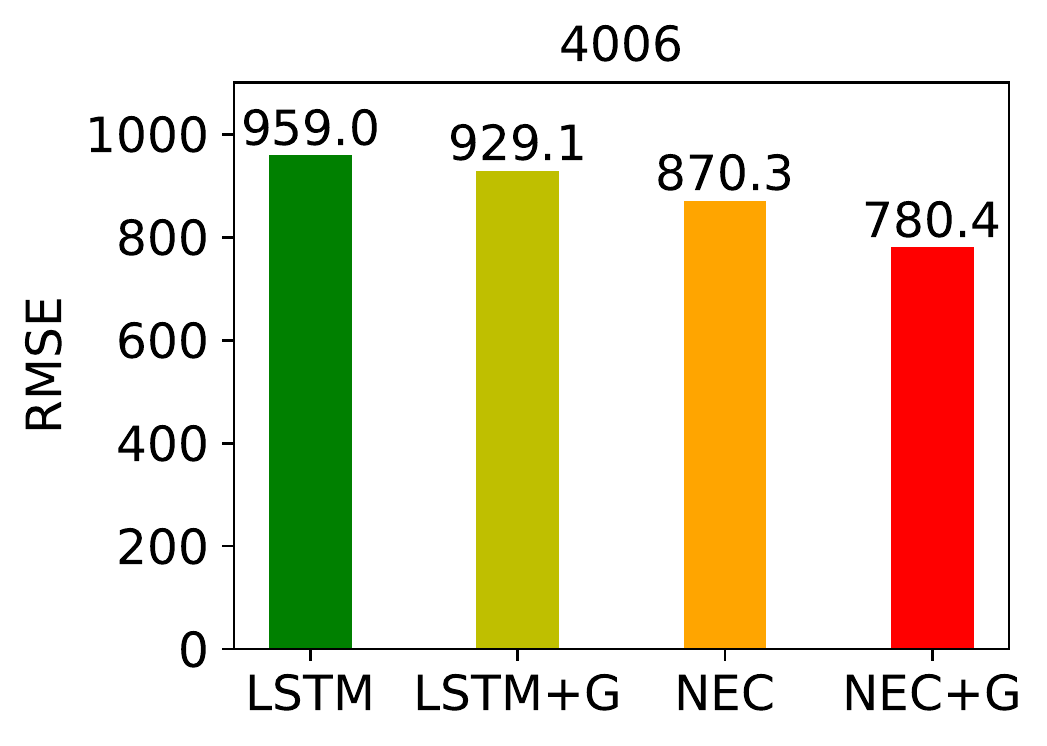}
\includegraphics[width=0.325\linewidth]{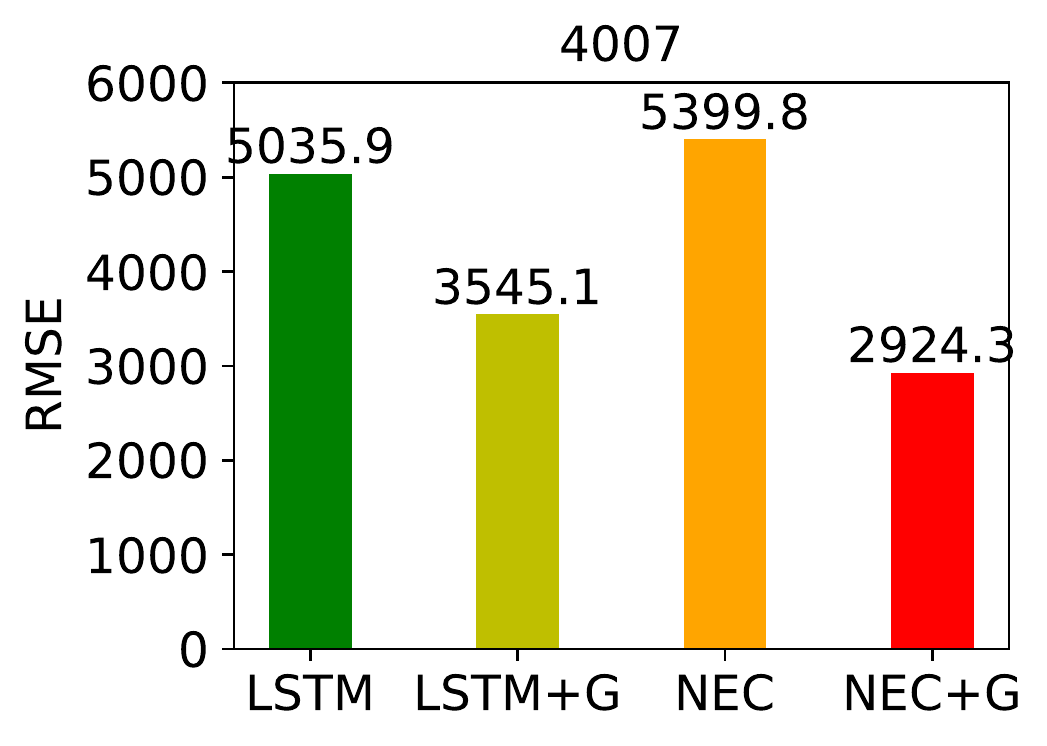}\\
\includegraphics[width=0.325\linewidth]{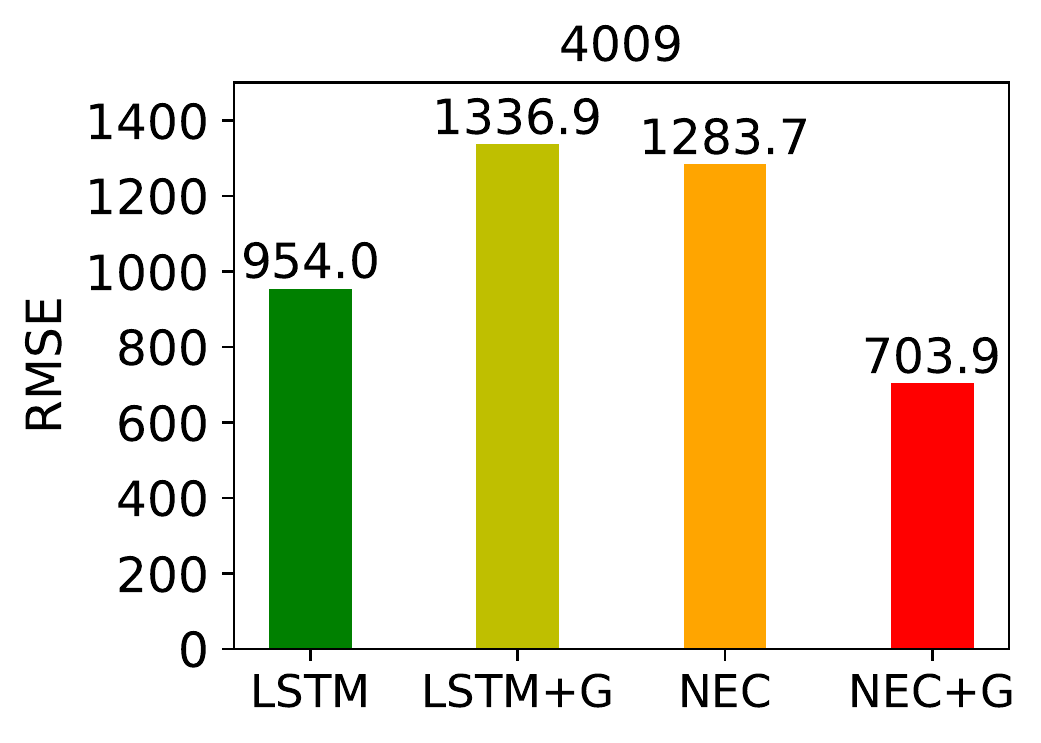}
\includegraphics[width=0.325\linewidth]{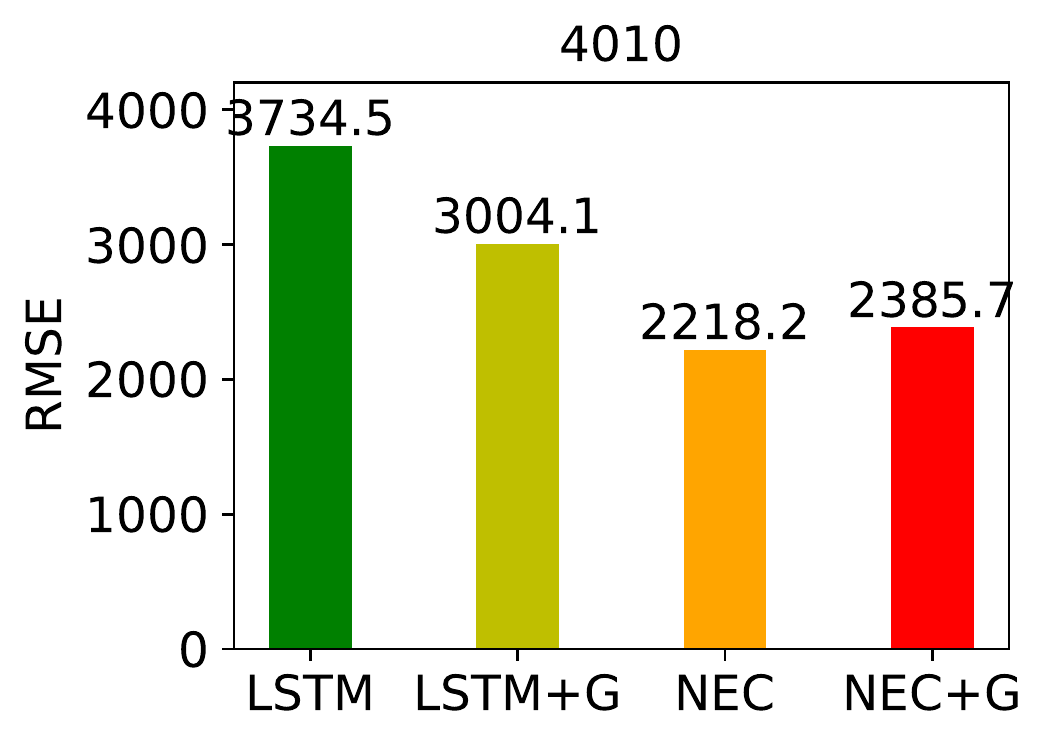}
\includegraphics[width=0.325\linewidth]{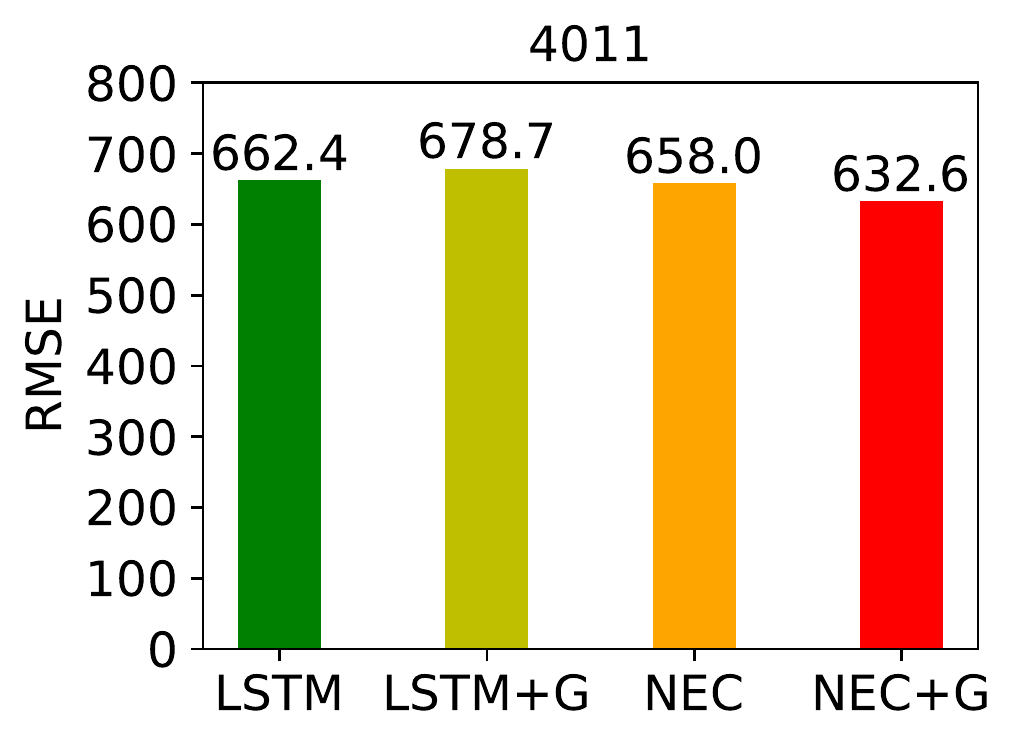}
\caption{Effectiveness comparison of NEC and LSTM variants with (LSTM+G and NEC+G) and without (LSTM and NEC) GMM indicator variables across all 9 sensors.}\label{fig:gmm-test}
\end{figure}
In the methods section, we hypothesized that adding the GMM indicator variable as an additional input to our model would help the component N and E models better distinguish between \textit{normal} and \textit{extreme} values. In order to test this hypothesis, we added GMM indicators to both the baseline LSTM model and our basic NEC model, and compared the test set effectiveness (RMSE). Figure~\ref{fig:gmm-test} shows the result of this experiment for our 9 sensors. In the figure, LSTM+G and NEC+G are the LSTM and NEC models with GMM indicator variables, respectively. Interestingly, adding the GMM indicator seems detrimental for the baseline LSTM, producing much worse RMSE scores for sensors 4009, and more than 5\% RMSE improvements only in sensors 4001, 4007, and 4010. The average RMSE improvement of the LSTM model when adding the GMM indicator across the 9 sensors is only 4\%. On the other hand, adding the GMM indicator produces significantly better results in 8 of the 9 sensors for our NEC model, and only slightly worse results for sensors 4010. The average RMSE improvement for NEC is 18\%, which is significantly better than the 4\% average RMSE improvement of the LSTM model. This shows that, while the GMM indicator may provide some limited benefit to a one-shot model like LSTM, it plays a much more important role in the success of our NEC+ framework.

\begin{table*}[htb]
\caption{Effectiveness Comparison (RMSE) of NEC+ Against Baselines for 9 Reservoirs}\label{tbl:effectiveness}
\centering
\footnotesize
\setlength{\tabcolsep}{2mm}{}
\begin{tabular}{lrrrrrrrrr}
\hline
% \multicolumn{1}{c}{\multirow{2}{*}{\textbf{Model}}} & \multicolumn{9}{c}{\textbf{Reservoir}}\\ 
% \cline{2-10} 
\textbf{Model/Reservoir} & \multicolumn{1}{c}{\textbf{4001}} & 
\multicolumn{1}{c}{\textbf{4003}} & \multicolumn{1}{c}{\textbf{4004}} & \multicolumn{1}{c}{\textbf{4005}} & \multicolumn{1}{c}{\textbf{4006}} & \multicolumn{1}{c}{\textbf{4007}} & \multicolumn{1}{c}{\textbf{4009}} & \multicolumn{1}{c}{\textbf{4010}} & \multicolumn{1}{c}{\textbf{4011}} \\ 
\hline
%  & \multicolumn{9}{c}{\textbf{RMSE}} \\
%   \cline{2-10} 
\textbf{ARIMA}& 1016.32 & 1859.70 & 2501.97 & 9692.87 & 1039.38 & 5854.48 & 1060.05 & 3465.20& 690.23\\ 
\textbf{Prophet} & 8469.74 & 38827.22 & 95279.31 & 181607.50 & 20904.57 & 187603.80 & 28629.44 & 114115.4 & 2829.26 \\
\textbf{LSTM} & 1167.73  & 1514.90  & 2342.71 & 6730.93  & 959.05  & 5035.91  & 954.04 & 3734.53 & 662.48  \\
\textbf{DNN-U}  & 1162.01  & 1597.72  & 3989.20 & 9878.41 & 983.27  & 4320.40  & 1411.63 & 4257.58  & 763.73  \\
\textbf{A-LSTM}  & 878.71  & 1536.04  & 2548.56 & 8919.33 & 1638.65  & 13529.86  & 1064.15 & 2914.75  & 700.50  \\
\textbf{N-BEATS}  & 937.24  & 1926.74  & 2280.83 & 7153.82 & 960.42  & 3153.76  & 1295.90 & 3162.17  & \textbf{514.30}  \\
\textbf{NEC+} & \textbf{740.19} & \textbf{1411.44}  & \textbf{1783.92} & \textbf{4352.74} & \textbf{780.46}  & \textbf{2092.73}  & \textbf{703.93} & \textbf{2275.48}  & 632.61  \\
% \hline
%  & \multicolumn{9}{c}{\textbf{MAPE}} \\
%   \cline{2-10} 
% \textbf{ARIMA}& 1.3573 & 0.7626 & 0.8694 & 1.2560 & 1.5401 & 0.8517 & 0.9504 & 1.7871 & 3.2914\\
% \textbf{Prophet} & 16.7877 & 19.8559 & 38.9642 & 35.6662 & 56.0537 & 32.9152 & 31.8069 & 45.2579 & 15.3312 \\
% \textbf{LSTM} & 1.6697  & 0.6153  & 0.7450 & 1.0092  & 1.3264  & 0.9253  & 0.9298 & 2.5520 & 3.1282  \\
% \textbf{DNN-U}  & 1.6509  & 0.6812  & 1.8738 & 1.9394 & 1.4551  & 0.6509  & 1.5604 & 2.1582  & 3.7131  \\
% \textbf{A-LSTM}  & 1.3533  & 0.6506  & 0.8424 & 1.2060 & 2.8017  & 2.1738  & 0.9705 & 1.3986  & 3.4137  \\
% \textbf{N-BEATS}  & 1.3346  & 0.7972  & 0.7882 & 1.1405 & 2.0061  & 0.4709  & 1.4580 & 1.7146  & \textbf{2.3108}  \\
% \textbf{NEC+} & \textbf{1.0319} & \textbf{0.5687}  & \textbf{0.6811} & \textbf{0.6350} & \textbf{1.0662}  & \textbf{0.3316}  & \textbf{0.5992} & \textbf{1.2894}  & 2.9237  \\
\hline
\end{tabular}
\end{table*}

\subsection{Effect of Adding Exogenous Variables}
\begin{table}[htb]
\centering
%\small
\footnotesize
\setlength{\tabcolsep}{2mm}{}
\caption{Effectiveness With/Without Exogenous Variables}\label{tbl:shed_rmse}
\begin{tabular}{lrrr}
\hline
% \multirow{2}{*}{\textbf{Model}}&\multicolumn{3}{c}{\textbf{Reservoir}}\\
% \cline{2-4}
\textbf{Model/Reservoir} & \multicolumn{1}{c}{\textbf{4005}} & \multicolumn{1}{c}{\textbf{4007}} & \multicolumn{1}{c}{\textbf{4010}} \\
\hline
LSTM & 6730.93 & 5035.91 & 3734.53 \\
LSTM+W & {\color{red}7568.68} & {\color{red}5728.30} & {\color{red}4145.16} \\
LSTM+G  & 6455.90 & 3545.19 & 3004.14  \\
LSTM+G+W & {\color{red}9760.62} & {\color{red}4128.37} & {\color{fgreen}2602.58} \\
NEC+G & 5114.49 & 2924.30 & 2385.77  \\
NEC+G+W (NEC+) & {\color{fgreen}\textbf{4352.74}} & {\color{fgreen}\textbf{2092.73}} & {\color{fgreen}\textbf{2275.48}} \\
\hline
\end{tabular}
\end{table}
An additional benefit may be obtained in NEC+ by including exogenous variables that may provide an additional signal that the model may use to learn the proper prediction function. In our experiments, we used watershed rain gauge time series data from the same times as our primary reservoir water level data to enhance models for reservoirs 4005, 4007, and 4010, as described in the methods section. We compared our NEC+ model with (NEC+G+W) and without (NEC+G) the watershed variables against variations of the baseline LSTM model with (LSTM+W, LSTM+G+W) and without (LSTM, LSTM+G) those same variables. The letter G in all model names indicates the presence of the GMM indicator variable. 

Table~\ref{tbl:shed_rmse} shows the results of our analysis. The watershed model results are colored green if they are better (lower RMSE) than the same model without watershed variables, and red if worse. We use bold to denote the best results across all models. Interestingly, including the watershed variables in the baseline LSTM and LSTM+G models leads to significantly worse results in most cases, but significantly better results (5\%--28\% lower RMSE) in the case of the NEC+G model. Our model can benefit more by focusing on normal or extreme prediction individually.

\subsection{Effect of Loss Function Parameters}

\begin{figure}[htb]
\centering
\includegraphics[width=0.5\linewidth]{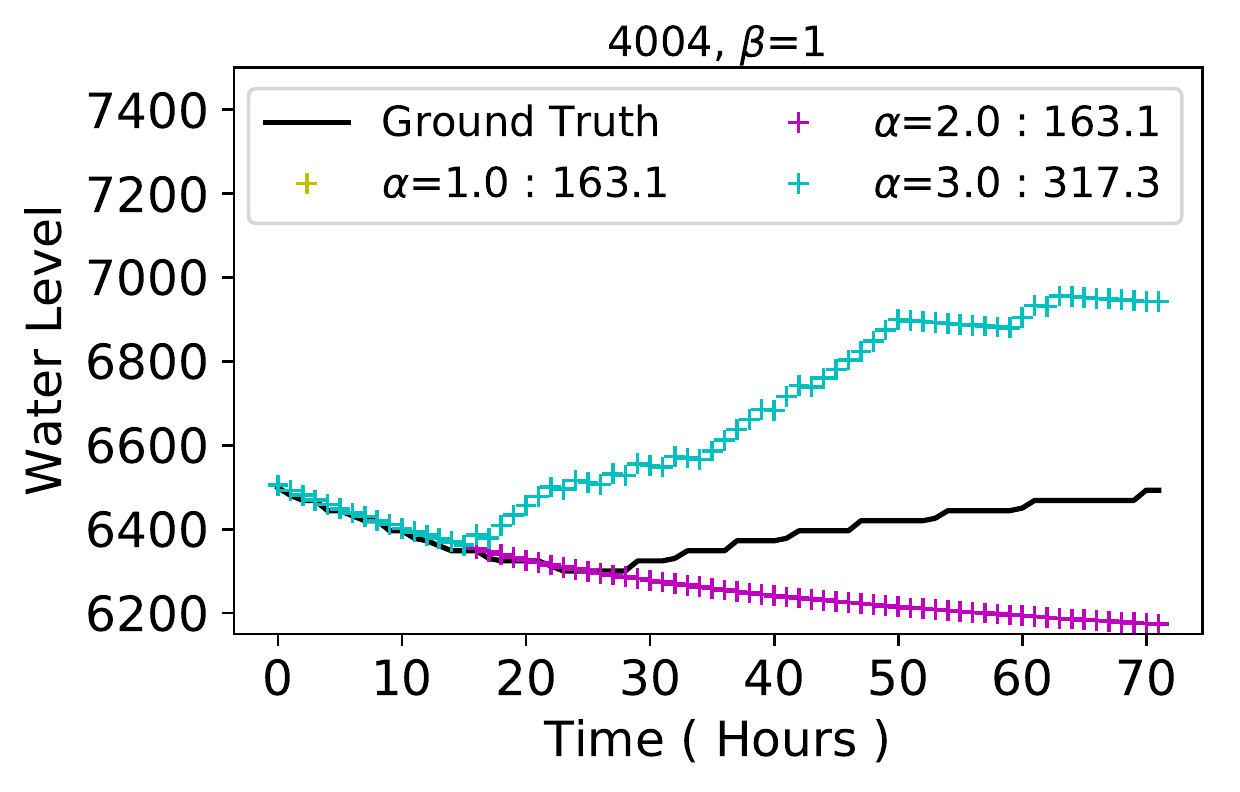}%\quad
\includegraphics[width=0.5\linewidth]{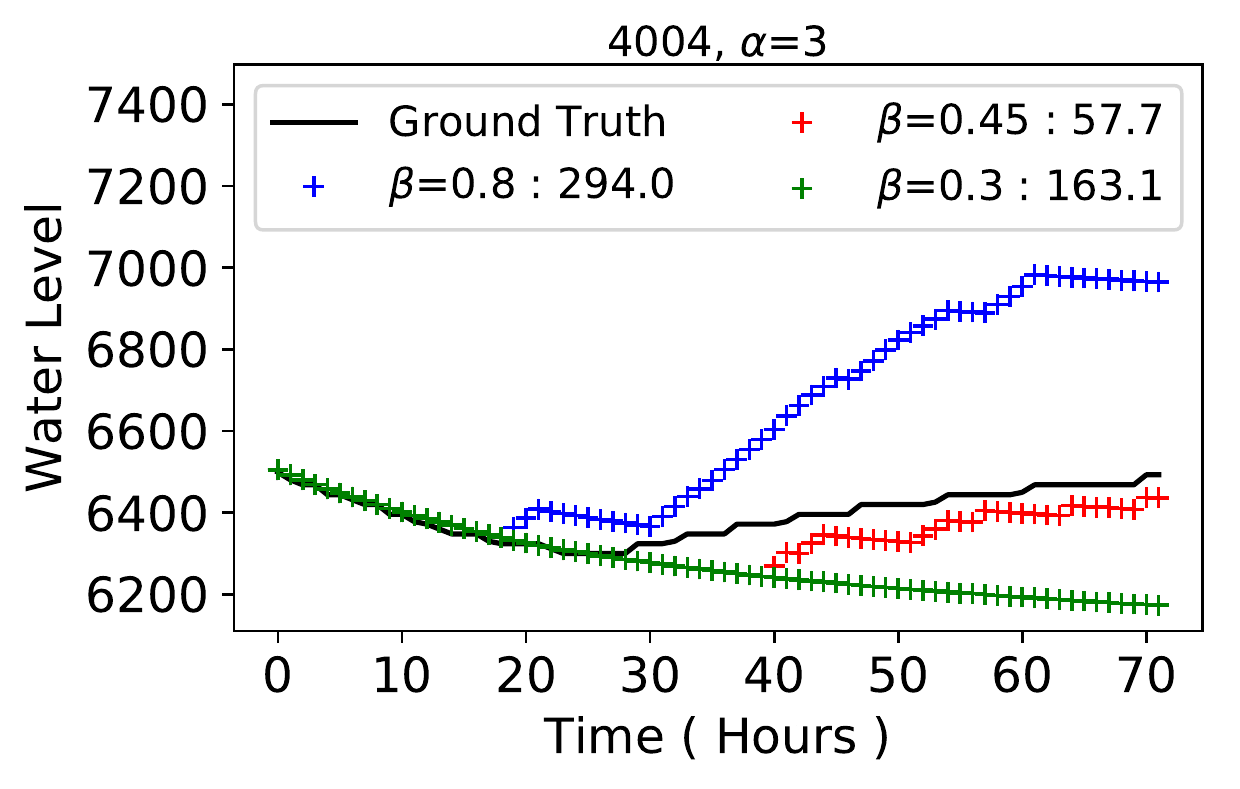}
\caption{Loss function parameter choices for sensor 4004.}\label{fig:loss}
\end{figure}

We proposed a parameterized loss function that we hypothesized would help improve the ability of our C model to pick out the rare extreme values and, as a result, lead to better prediction of future values. As a way to see how the two parameters of our loss function may affect the prediction, we trained several models with different values of $\alpha$ (the BCE power) while keeping $\beta$ (BCE vs. RMSE strength) constant, and several with varying $\beta$ while keeping $\alpha$ constant, the results of which can be seen in \figurename~\ref{fig:loss}. As expected, increasing the $\alpha$ parameter (top figure) leads to more values being classified as \textit{extreme}, allowing the E model to play a bigger role in the NEC+ model. When $\alpha$ if too large (bottom figure), its effect can be dampened by decreasing the value of $\beta$. Therefore, we suggest keeping $\beta=1$ while tuning $\alpha$ and then tuning $\beta$ for the best found $\alpha$.

\subsection{Effectiveness of NEC+ Against Baselines}

\begin{figure}[tbh]
    \centering
    \includegraphics[width=\linewidth]{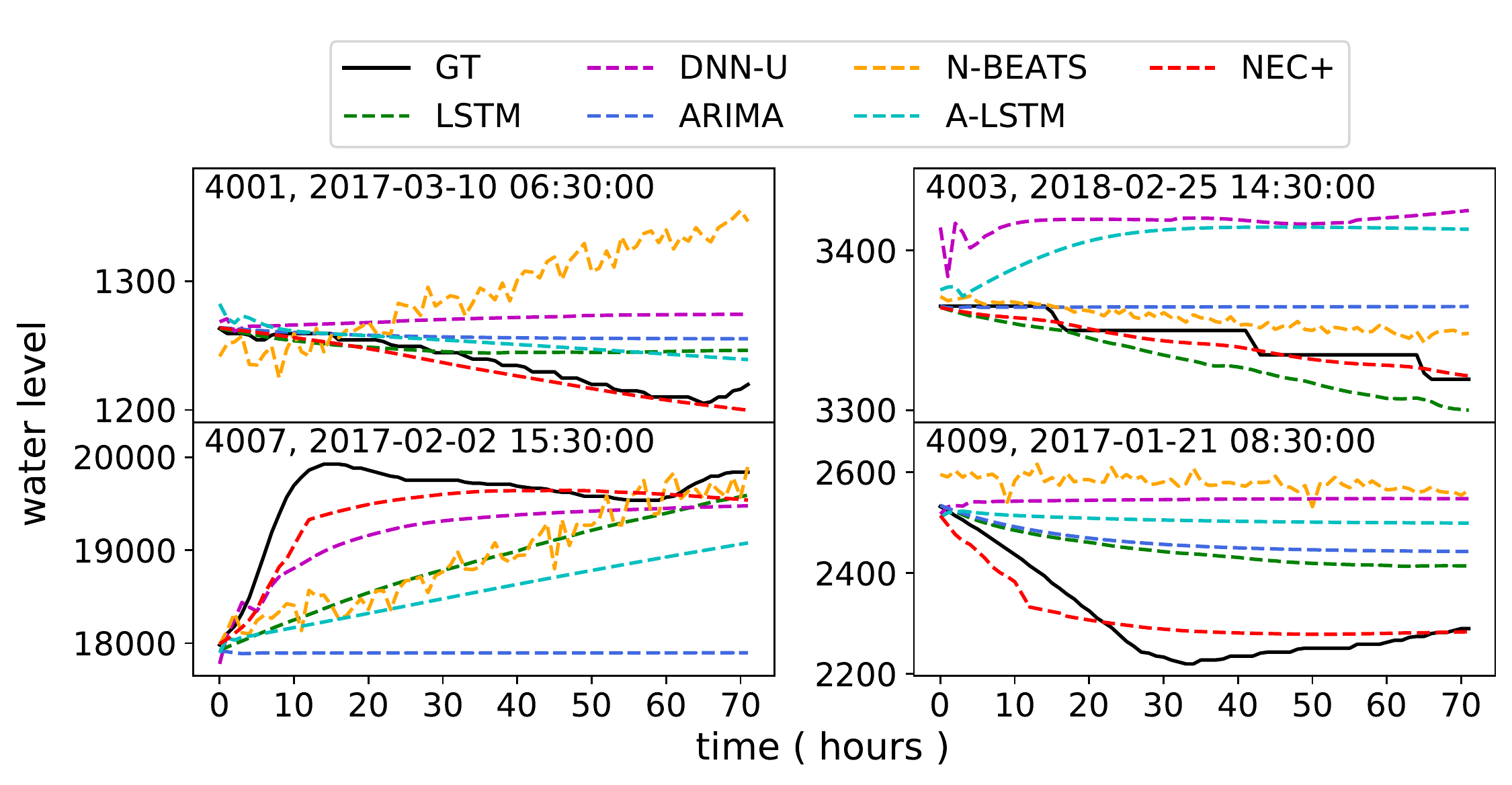}
    \caption{Example 3 days ahead predictions for four sensors.}
    \label{fig:predictions}
\end{figure}

% \begin{figure*}[htb]
% \captionsetup[subfloat]{farskip=3pt,captionskip=2pt}
% \centering
% \subfloat{
%   \includegraphics[width=0.475\textwidth]{4001_2_v2.pdf}
% \label{S4001-3}}%
% \subfloat{
%   \includegraphics[width=0.475\textwidth]{4001_3_v2.pdf}
% \label{R4005-2}}\\
% \subfloat{
%   \includegraphics[width=0.475\textwidth]{4001_5_v2.pdf}
% \label{R4005-10}}
% \subfloat{
%   \includegraphics[width=0.475\textwidth]{4005_26_v2.pdf}
% \label{R4007-2}}\\%
% \subfloat{
%   \includegraphics[width=0.475\textwidth]{4007_2_v2.pdf}
% \label{R4007-3}}%
% \subfloat{
%   \includegraphics[width=0.475\textwidth]{4007_3_v2.pdf}
% \label{R4007-6}}%
% \caption{Example 3 days ahead predictions for regions with \textit{extreme} changes (\textbf{a}--\textbf{f}) and \textit{normal} changes (\textbf{g}--\textbf{i}). The title of each sub-figure shows the start timestamp of the prediction and the legend includes RMSE values for each method.}\label{fig:effectiveness}
% \end{figure*}

Our main evaluation question was whether our proposed NEC+ model is an effective method for solving the 3-day ahead prediction problem in time series with \textit{extreme} events such as our 9 reservoirs. To answer this question, we compared NEC+ against a variety of traditional and state-of-the-art methods and Table~\ref{tbl:effectiveness} presents the RMSE results of all these models. Equivalent MAPE values are included in Table~\ref{tbl:mape} in the appendix. Values in bold are the best (lowest) RMSE for each sensor. 
Our model significantly outperforms all traditional methods (ARIMA, Prophet, and LSTM) and state-of-the-art methods DNN-U and A-LSTM for all 9 sensors. However, the results for NEC+ are, on average, 37\% and 33\% better than those of DNN-U and A-LSTM, respectively, across all 9 sensors. Moreover, DNN-U and A-LSTM were unable to outperform the traditional ARIMA or LSTM baselines for 6 and 3 out of the 9 sensors, respectively, pointing to their overall instability. The N-BEATS model was the most competitive, outperforming NEC+ on only one sensor, 4011. However, NEC+ results are significantly better than those of N-BEATS (Wilcoxon T test s=1, p=0.0078).

\figurename~\ref{fig:predictions} shows some example 3-day predictions from our test set for four of the sensors (due to lack of space). Predicted time series for other sensors are included in the technical appendix. We did not include Prophet in the results as the model performed very poorly and would impede visualizing the performance of the remaining models. Overall, NEC+ is able to more closely predict the ground truth water level values, both in the presence of \textit{extreme} events and during normal conditions. ARIMA often misses the mark and DNN-U, LSTM, A-LSTM, and N-BEATS sometimes follow the trend of the ground truth and sometimes do not. Overall, NEC+ shows it can more closely account for extreme changes in the time series.

% The oversampling ratio for the E model and Classifier datasets is set to 0, indicating that at least one extreme value exists in the predicted part of a training sample. This is natural for the E model because only those extreme values are used in the loss and gradient computation. Back propagation will not occur in a sample with only normal values. Furthermore, because we are forecasting the classifying labels when extreme is the minority class, oversampling is used in this circumstance for the Classifier. This choice is supported by the grid search results in Classifier model training.

% \subsubsection{\textbf{Influence of GMM algorithm}}
% In Figure 7, the axis x is the input of GMM and axis y is output, which shows that for different sensors, the bound of \textit{normal} and \textit{extreme} indicator changes a lot. When the bound is similar to 1.5 which we use as threshold in NEC+, as sensor 4005 and 4007, the better prediction performance achieves. As for sensor 4011, the GMM indicator classify those values within $5*std$ as normal ones, NEC+ cannot get benefits from it. One direction is to improve the threshold in NEC+, but this can cause more imbalance between \textit{normal} and \textit{extreme} value, so give the classifier more challenge. The other way is to tune the components of GMM model so change the unlabeled clustering results to get a new indicator.

\section*{Conclusion}

In this work, we presented a novel composite framework and model, NEC+, designed to better account for rare yet important extreme events in long single- and multi-variate time series. Our framework learns distinct regression models for predicting \textit{extreme} and \textit{normal} values, along with a merging classifier that is used to choose the appropriate model for each future event prediction. NEC+ uses an unsupervised clustering approach to dynamically produce distribution indicators, which improves the model's robustness to the occurrence of severe events. In addition, to improve training performance, our framework uses a selected backpropagation approach and a two-level sampling algorithm to accommodate imbalanced extreme data. A parameterized loss function is also proposed to improve the NEC+ classifier performance.
Extensive experiments using more than 31 years of reservoir water level data from Santa Clara County, CA, showed that the components of the NEC+ framework are beneficial towards improving its performance and that NEC+ provided significantly better predictions than state-of-the-art baselines (Wilcoxon T test p-values between 0.0039 and 0.0078).

%%%%%%%%%%%%%%%%%%%%%%%%%%%%%%%%%%%%%%%%%%%%%%%%%%%%%%%%%%%%%%%%%%%%%%%%%%%%%%%%

%%%%%%%%%%%%%%%%%%%%%%%%%%%%%%%%%%%%%%%%%%%%%%%%%%%%%%%%%%%%%%%%%%%%%%%%%%%%%%%%

%%%%%%%%%%%%%%%%%%%%%%%%%%%%%%%%%%%%%%%%%%%%%%%%%%%%%%%%%%%%%%%%%%%%%%%%%%%%%%%%
\section*{Appendix}

\section{Additional Related Work}\label{sec:related2}
In this section, we detail additional related work employing traditional machine learning methods for the task of water level prediction. Castillo-Bot{\'o}n et al.~\cite{CastilloBotn2020AnalysisAP} applied support vector regression (SVR), Gaussian processes, and artificial neural networks (ANNs) to obtain short-term water level predictions in a hydroelectric dam reservoir. Several studies by Nayak et al.~\cite{NAYAK200452}, Larrea et al.~\cite{w13152011}, and Tsao et al.~\cite{en14123461} use neural network models (NN) and adaptive neuro-fuzzy inference systems (ANFIS)~\cite{Chang2006AdaptiveNI} to forecast the water level. Some machine learning algorithms, such as Gaussian process regression (GPR) and quantile regression, can not only predict but also quantify forecast uncertainty. Tree-based models are computationally inexpensive and have the advantage that they do not assume any specific distribution in the predictors~\cite{RFbased}. Classification and regression trees (CARTs) and random forest (RF) have also been used to solve hydrologic prediction problems~~\cite{RFbased}. Nguyen et al.~\cite{Nguyen_2021} proposed a hybrid model for hourly water level prediction that integrates the XGBoost model with two evolutionary algorithms and showed that it outperformed RF and CART in the multistep-ahead prediction of water levels. While neural networks were used in some works for water level prediction, they were usually shallow networks that were not able to recognize complex patterns in the data, so feature engineering and extensive manual tuning based on domain expertise had to be employed to improve their performance~\cite{Li01}. 

\begin{table}[tbh]
\caption{Reservoir and Rain Sensor Data}\label{tbl:data}
\footnotesize
\setlength{\tabcolsep}{2mm}{}
\begin{tabular}{ccccc}
\textbf{ID} & \textbf{Type} & \textbf{Name/Location}  & \textbf{Start} & \textbf{End} \\ \hline
\textbf{4001} & Reservoir & Almaden   & 10/1973 & 06/2020  \\
\textbf{4003} & Reservoir & Calero    & 06/1974 & 06/2020  \\
\textbf{4004} & Reservoir & Chesbro   & 02/1974  & 06/2020  \\
\textbf{4005} & Reservoir & Coyote    & 10/1973 & 06/2020  \\
\textbf{4006} & Reservoir & Guadalupe & 10/1973 & 06/2020  \\
\textbf{4007} & Reservoir & Lexington & 10/1973 & 06/2020  \\
\textbf{4009} & Reservoir & Stevens Creek & 10/1973 & 06/2020 \\
\textbf{4010} & Reservoir & Uvas    & 10/1973 & 06/2020  \\
\textbf{4011} & Reservoir & Vasona  & 11/1973 & 06/2020  \\
\textbf{6017} & Rain & Coe Park & 02/1980 & 07/2019  \\
\textbf{6135} & Rain & Uvas Canyon Park & 07/1991 & 07/2019  \\
\textbf{6044} & Rain & Lome Prieta & 07/1989   & 07/2019   \\
\textbf{6069} & Rain & Mt Umunhum & 07/1989   & 07/2019     
\end{tabular}
\end{table}

\begin{figure}[b]
 \centering
 \includegraphics[width=0.31\textwidth]{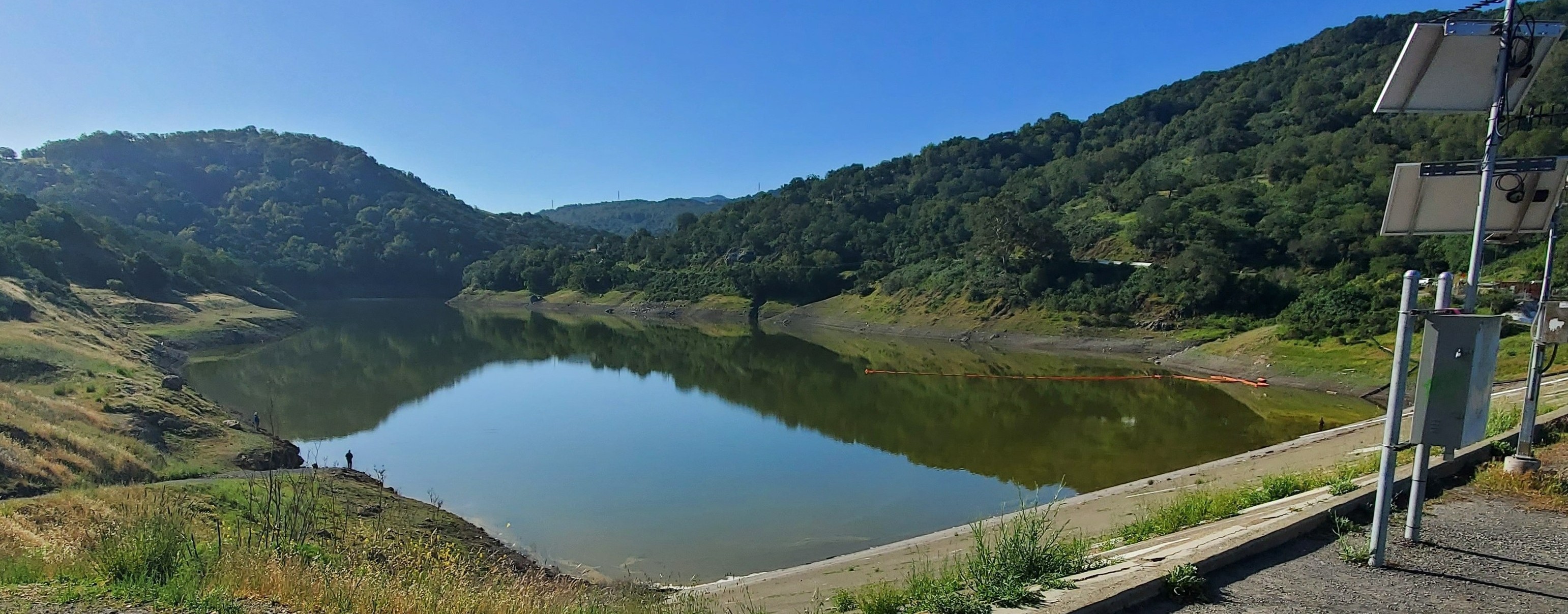}
  \caption{Guadalupe Reservoir, CA.}\label{fig:reservoir}
\end{figure}

\section{Data and Preprocessing}
Table~\ref{tbl:data} shows the 9 reservoir sensors and 4 rain sensors used in our study, along with their location and period of time we had data for. As an illustration, \figurename~\ref{fig:reservoir} shows Guadalupe Reservoir (ID 4006). Information about these reservoirs can be found at \url{https://www.valleywater.org/your-water/local-dams-and-reservoirs}. While data for most reservoir sensors was available from 1974, we observed many missing and outlier data points in early years of the series due to sensor or data storage failures. Therefore, we limited our analysis to the years 1988--2019 for univariate models involving only reservoir sensor data, and to the years 1991--2019 for multivariate models using both reservoir and rain sensor data. Short gaps in the time series during these periods were filled in via an adaptive polynomial interpolation approach that learned a polynomial function to best fit $k$ values before and after a gap of size $2k$ and projected the missing values onto that function. A dataset including relevant data used in this study will be made available to researchers upon publication of this paper.

\begin{figure*}[htb]
 \centering
 \scriptsize
  \includegraphics[width=1\textwidth]{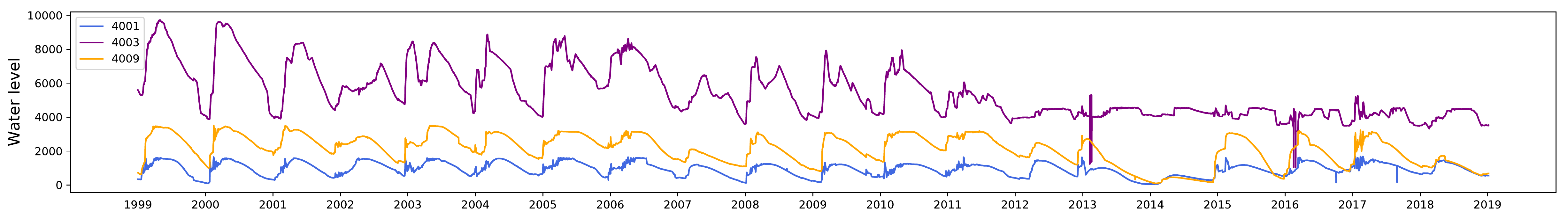}\\
  Year\\
  \includegraphics[width=0.33\textwidth]{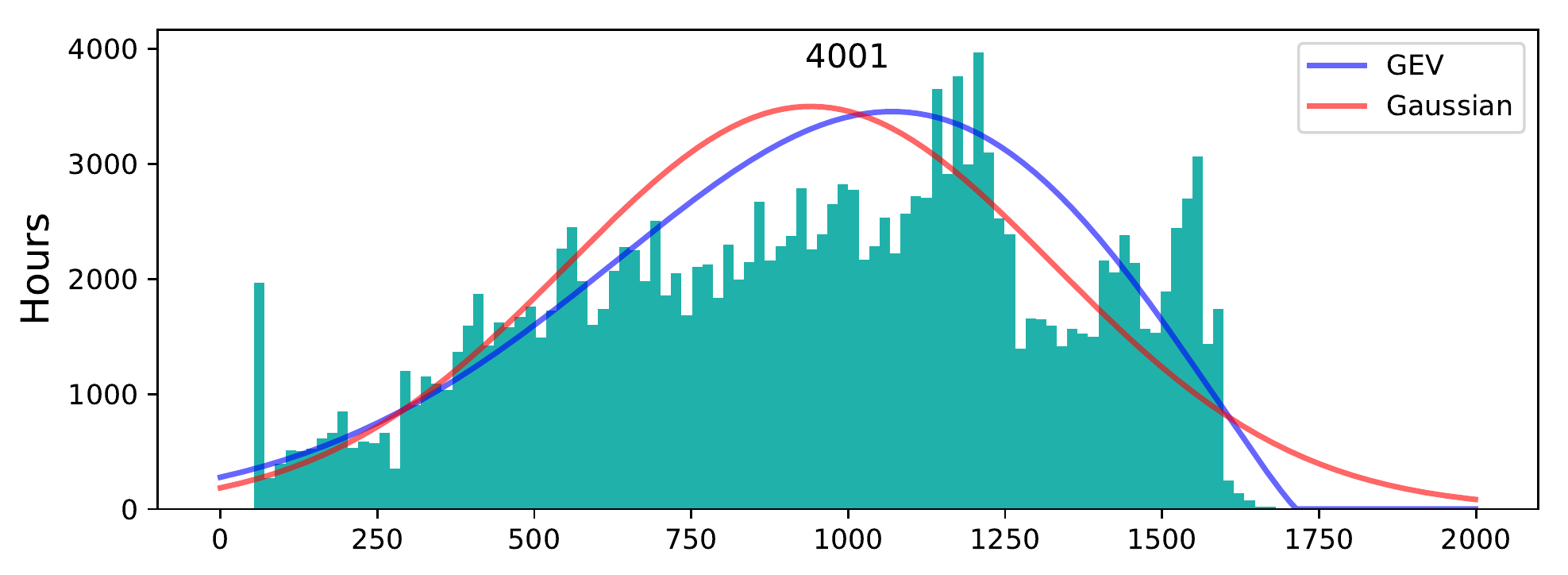}
  \includegraphics[width=0.33\textwidth]{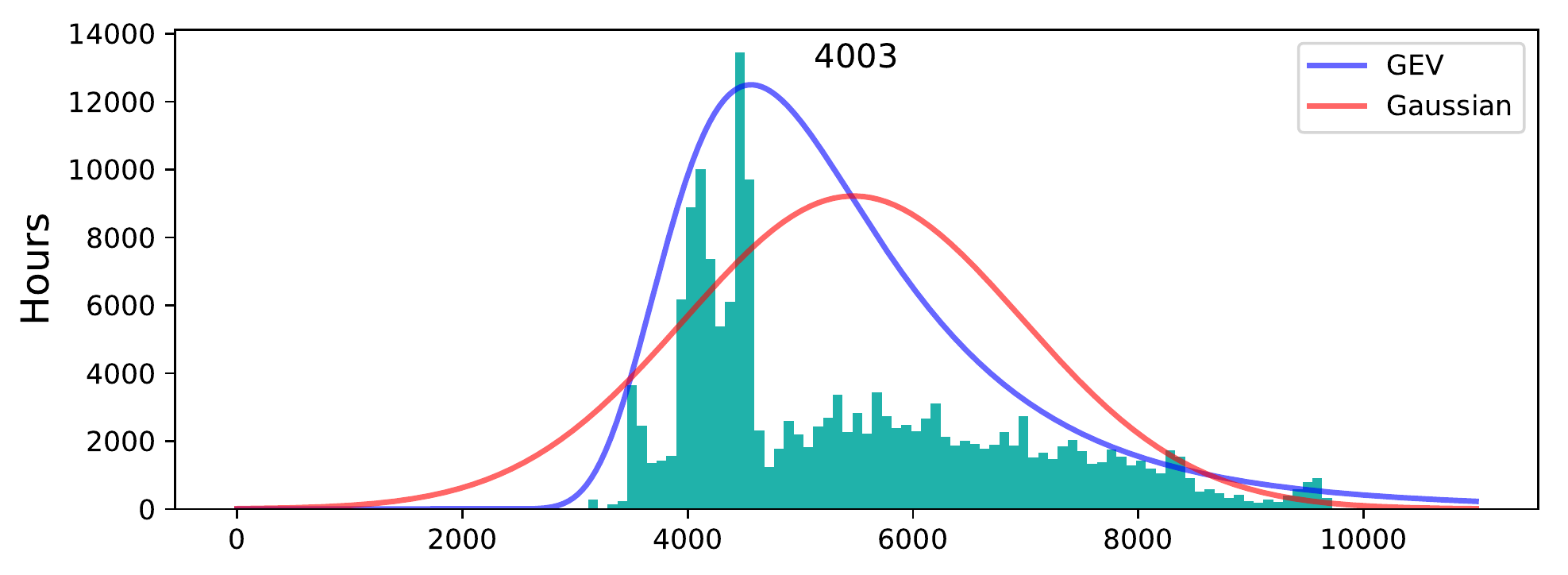}
  \includegraphics[width=0.33\textwidth]{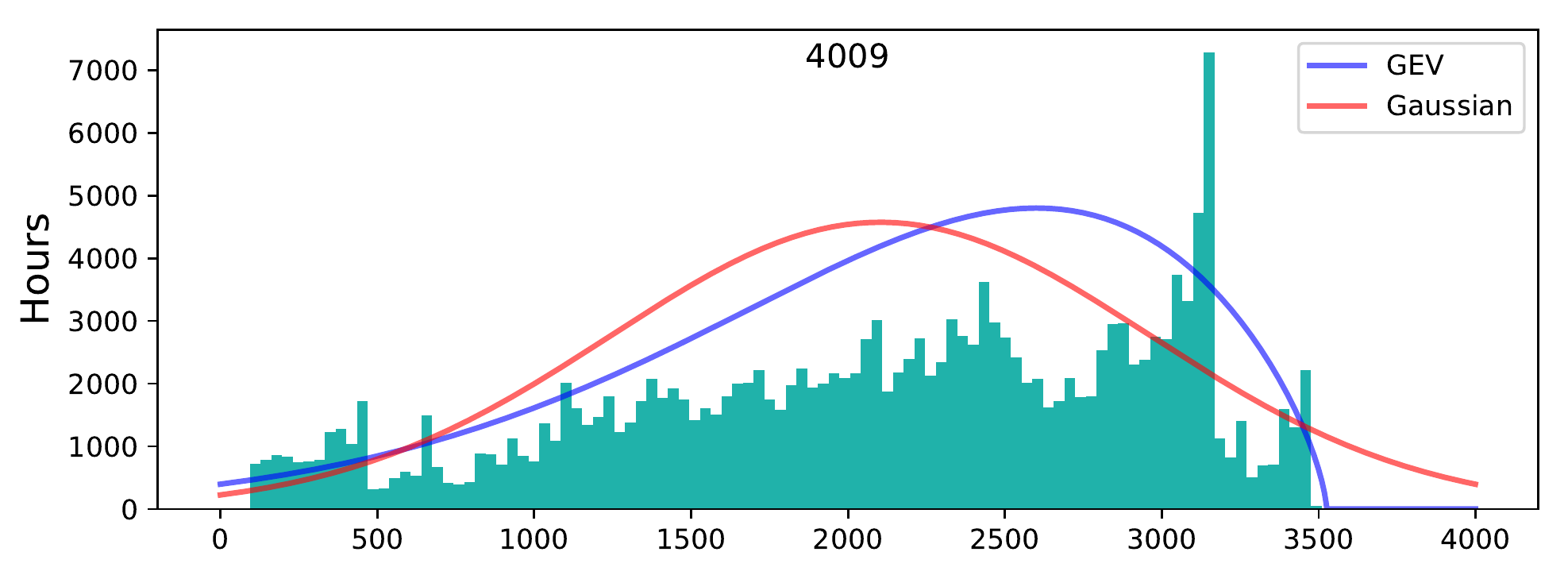}\\
  Water Level\\
  \includegraphics[width=0.33\textwidth]{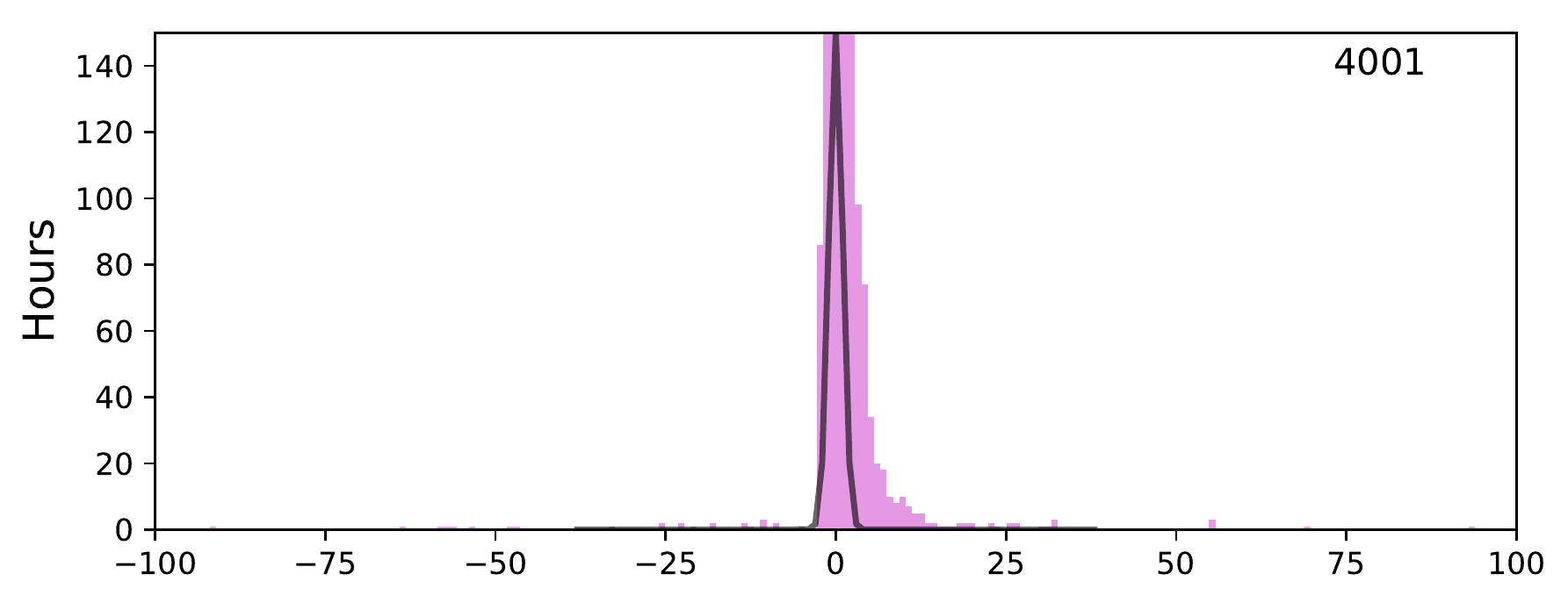}
  \includegraphics[width=0.33\textwidth]{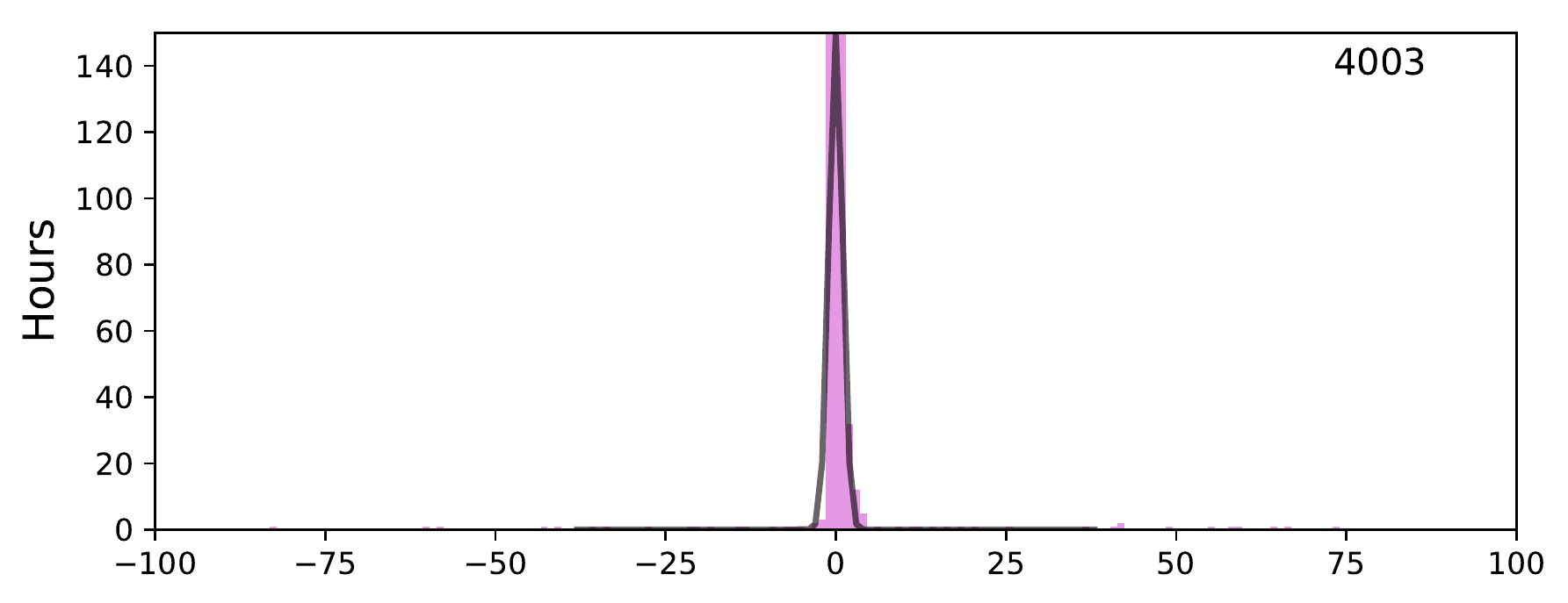}
  \includegraphics[width=0.33\textwidth]{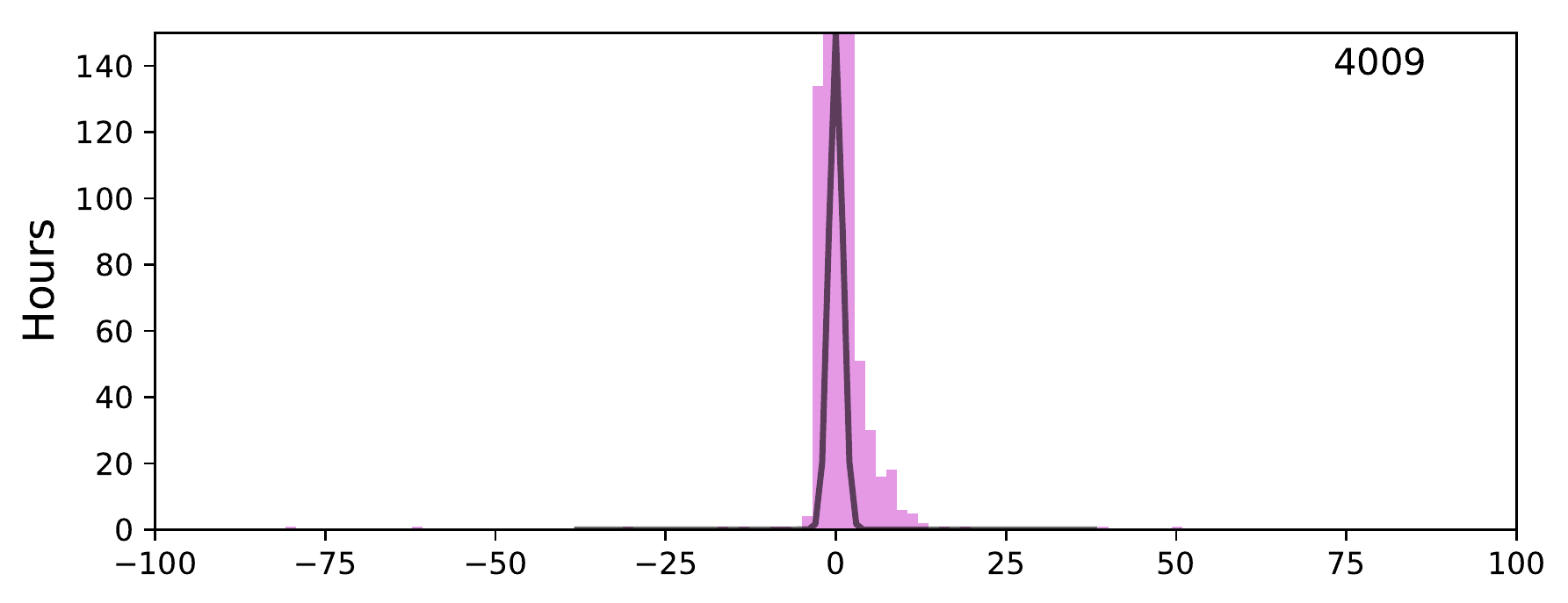}\\
  Standardized First-Order Difference of the Water Level
  \caption{Water level values (top), extreme value distribution pdf fit of water levels (middle), and Gaussian distribution pdf fit of water level standardized first difference (bottom) for 3 reservoirs across 20 years.}\label{fig:wave2}
\end{figure*}

\section{Extreme Events}

Figure~\ref{fig:wave2} shows water levels for three of our 9 sensors across a period of 20 years (top) and GEV and Gaussian probability density functions (pdf) we fitted for the same water level values (middle). As can be seen from the middle charts, the GEV distribution provides a better fit, showing the presence of extreme values in our data. In particular, the RMSE of the Gaussian distribution fit is 26.9\%, 46.0\%, and 37.2\% higher than that of the GEV distribution fit for the 4001, 4003, and 4009 reservoirs, respectively. 

We follow a standard time series analysis preprocessing approach and obtain a stationary time series by applying first-order differencing and then standardizing the resulting time series values, %$x_t' = x_t-x_{t-1}$, and $x' = \frac{x'-\mu}{\sigma}$,
\begin{align*}
x_t' &= x_t-x_{t-1},\\
x' &= \frac{x'-\mu}{\sigma},
\end{align*}
where $\mu$ and $\sigma$ here are the mean (location parameter) and standard deviation (scale parameter) of the Gaussian distribution of the time series $x'$. After obtaining predictions for a time series, we use the same location and scale parameters to inverse the standardization, and the last ground truth value in the time series to inverse the first-order differencing, obtaining values in the same range as the original time series. The bottom charts in Figure~\ref{fig:wave2} show the pdf of the differenced and standardized time series values for the same sensors as in the top and middle figures, along with the best fit Gaussian for those same values. The y-axis is limited to the range $[0,150]$ for visibility but otherwise stretches to $80,000$ for the 0 water level difference bar in the histograms of the three sensors (most of the time there is no change in water level from one hour to the next), resulting in very thin and tall Gaussian distributions.

\section{Sampling Policies}

\begin{table}[htbp]
\centering
\footnotesize
\setlength{\tabcolsep}{2mm}{}
\caption{Over Sampling for Sensor 4005}\label{tbl:over-sampling}
\begin{tabular}{cccccc}
\hline
\textbf{Models} & \textit{RMSE} & OS=0.3 & OS=0.2 & \textbf{OS=0.04} & OS=0 \\
\hline
\multirow{3}{*}{L+G} & Total & 22495.9 & 9558.5 & \textbf{7919.8} & 8208.4 \\
& Normal & 21788.4 & 8894.5 & 6872.9 & 7162.1 \\
& Extreme & 1992.1 & 1558.0  & 2474.4 & 2478.7 \\
\hline
\multirow{3}{*}{L} & Total & 68442.4 & 12300.9 & \textbf{7769.6} & 8183.0 \\
& Normal & 68170.3 & 11556.7 & 6784.9 & 7122.4 \\
& Extreme & 917.1 & 1932.6 & 2654.3 & 2512.0 \\
\hline
\end{tabular}
\end{table}

To showcase the importance of sampling in our work, we ran an experiment in which we trained an LSTM model and one that also used GMM features (L and L+G in Table~\ref{tbl:over-sampling}), and allowed different levels of oversampling; OS=0.04 means that 4\% of the training samples had at least one \textit{extreme} value in the prediction area of the sample, even though only 0.5\% of the values in sensor 4005 were deemed \textit{extreme}. Table~\ref{tbl:over-sampling} shows the total RMSE as well as component RMSE scores computed only for the \textit{normal} and \textit{extreme} values. While the extreme values RMSE continues to decrease as the OS level increases, oversampling lead to marginal improved total RMSE scores at OS=0.04 and markedly worse results for higher levels of oversampling. This shows that oversampling cannot be used as a panacea to account for the rarity of the \textit{extreme} values in the time series. Instead, our NEC framework separates the task of predicting \textit{extreme} and \textit{normal} values, achieving markedly improved results in the process.

% \section{Selected Backpropagation in the Normal and Extreme models}
% \begin{figure}[htb]
%  \centering
%   \includegraphics[width=0.4\textwidth]{computation_graph.pdf}
%   \caption{Computational graph for the N and E models. Blue inputs represent \textit{normal} values while orange ones represent \textit{extreme} values, which will not take part in the backpropagation computation in the N model.}\label{fig:backprop}
% \end{figure}
% \figurename~\ref{fig:backprop} represents visually the backpropagation mechanism across the computational graph of our normal (N) and extreme (E) models.

\begin{figure*}[htb]
 \centering
 \includegraphics[width=\textwidth]{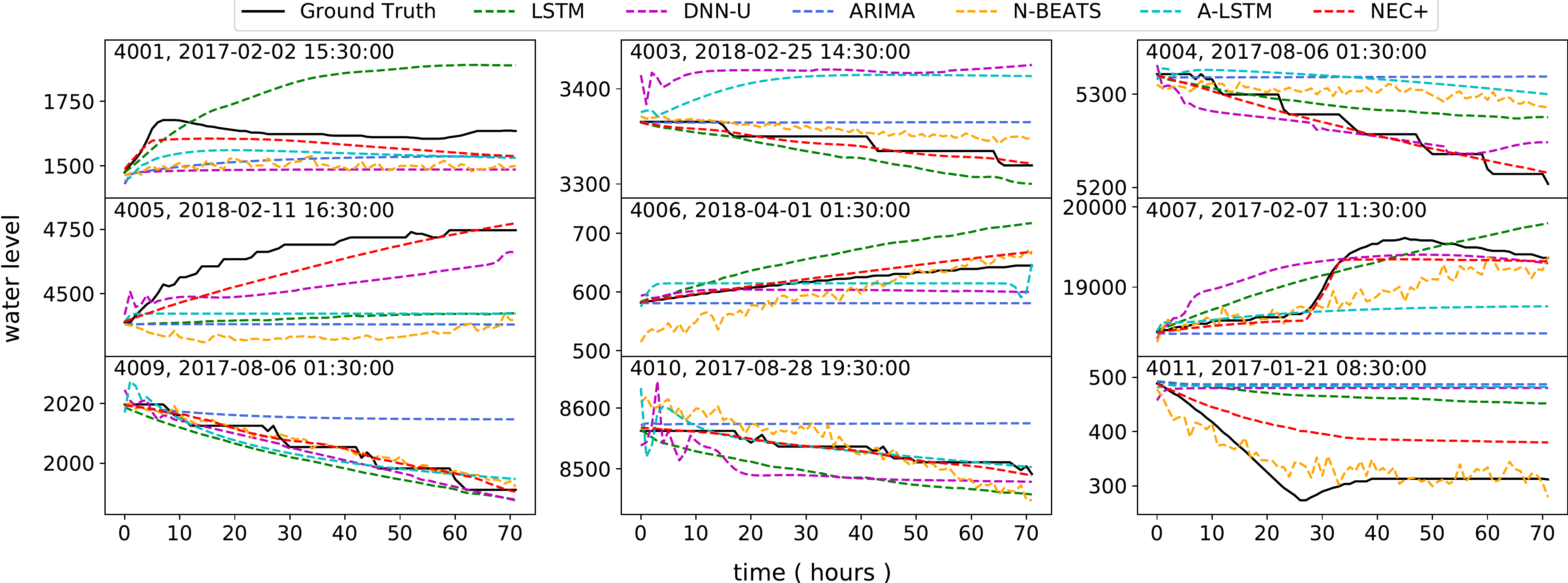}
  \caption{Example 3 days ahead predictions for each of our 9 sensors. Best viewed in color.\vspace{15pt}}\label{fig:predictions_all}
\end{figure*}

\begin{table*}[htb]
\caption{MAPE of NEC+ vs. Baselines for 9 Reservoirs}\label{tbl:mape}
\centering
\footnotesize
\setlength{\tabcolsep}{2mm}{}
\begin{tabular}{lrrrrrrrrr}
\hline
\textbf{Model/Reservoir} & \multicolumn{1}{c}{\textbf{4001}} & 
\multicolumn{1}{c}{\textbf{4003}} & \multicolumn{1}{c}{\textbf{4004}} & \multicolumn{1}{c}{\textbf{4005}} & \multicolumn{1}{c}{\textbf{4006}} & \multicolumn{1}{c}{\textbf{4007}} & \multicolumn{1}{c}{\textbf{4009}} & \multicolumn{1}{c}{\textbf{4010}} & \multicolumn{1}{c}{\textbf{4011}} \\ 
\hline
\textbf{ARIMA}& 1.3573 & 0.7626 & 0.8694 & 1.2560 & 1.5401 & 0.8517 & 0.9504 & 1.7871 & 3.2914\\
\textbf{Prophet} & 16.7877 & 19.8559 & 38.9642 & 35.6662 & 56.0537 & 32.9152 & 31.8069 & 45.2579 & 15.3312 \\
\textbf{LSTM} & 1.6697  & 0.6153  & 0.7450 & 1.0092  & 1.3264  & 0.9253  & 0.9298 & 2.5520 & 3.1282  \\
\textbf{DNN-U}  & 1.6509  & 0.6812  & 1.8738 & 1.9394 & 1.4551  & 0.6509  & 1.5604 & 2.1582  & 3.7131  \\
\textbf{A-LSTM}  & 1.3533  & 0.6506  & 0.8424 & 1.2060 & 2.8017  & 2.1738  & 0.9705 & 1.3986  & 3.4137  \\
\textbf{N-BEATS}  & 1.3346  & 0.7972  & 0.7882 & 1.1405 & 2.0061  & 0.4709  & 1.4580 & 1.7146  & \textbf{2.3108}  \\
\textbf{NEC+} & \textbf{1.0319} & \textbf{0.5687}  & \textbf{0.6030} & \textbf{0.6350} & \textbf{1.0662}  & \textbf{0.3316}  & \textbf{0.5992} & \textbf{1.2894}  & 2.9237  \\
\hline
\end{tabular}
\end{table*}

\section{Effectiveness Comparison}
Table~\ref{tbl:mape} shows MAPE values for the same test set predictions whose RMSE values are shown in Table~\ref{tbl:effectiveness} in the main paper. The best results for each sensor are displayed in bold. As can be seen, and as expected, the results mimic the ones in the main paper. Our method, NEC+ is able to outperform all standard and state-of-the-art baselines for all but one sensor.

\figurename~\ref{fig:predictions_all} provides additional example predictions, one for each of our 9 sensors, showcasing typical prediction results for NEC+ and its baselines. Results show that NEC+ predictions are able to more closely follow the ground truth values, showcasing its adaptability in the presence of extreme events.

\section{Metaparameter Choices}
In our study, we fixed $f$ to 72 forecasting points (3 days ahead prediction) and experimented with input length of history $h$ in [72, 120, 240, 360, 720], extreme threshold $\epsilon$ in [1.2, 1.5, 2.0], GMM indicator components $M$ in [2, 3, 4, 5] and oversampling ratio $OS\%$ in [0,  0.04, 0.2, 0.3, 0.5, 0.8, 1.0]. Table~\ref{tbl:metaparameters} shows the final parameters chosen for each sensor after our parameter study. We want to emphasize how similar the architecture and even the hyperparameters are across models for various sensor datasets in our experiments. This shows that the suggested architecture generalizes effectively to reservoirs with various distributions. The same design was successfully applied to other sensors after first being tuned for sensor 4005. We then performed a second round of tuning for sensor 4009 and successfully applied the learned parameters to some of the other sensors to improve their performance. %, demonstrating the improvement potential of NEC+. 
As a result, there are primarily two groups of parameter sets for the N, E, and C models, each with a different batch size, number of layers, and volume of training samples, as detailed in Table~\ref{tbl:metaparameters}. We used early stopping in our training with a threshold of 3 iterations for the N model and 4 iterations for the E and C models. %Compared to those who had to grid search for each dataset separately, or only verify the architecture on one hydro dataset~\cite{w14010034,le2021attention}, this is a substantially stronger outcome.

\begin{table*}[htb]
\caption{Metaparameters}\label{tbl:metaparameters}
\centering
\footnotesize
\begin{tabular}{lrrrrrrrrrr}
\hline
\textbf{Meta-parameter} & \textbf{4001} & \textbf{4003} & \textbf{4004} & \textbf{4005} & \textbf{4006} & \textbf{4007} & \textbf{4009} & \textbf{4010} & \textbf{4011}               \\
\hline
Input length $h$ & 360 & 360 & 360 & 360 & 360 & 360 & 360 & 360 & 360    \\
Extreme threshold $\epsilon$ & 1.5 & 1.5 & 1.5 & 1.5 & 1.5 & 1.5 & 1.8 & 1.5 & 1.5    \\
GMM components $M$ & 3 & 3 & 3 & 3 & 3 & 3 & 4 & 3 & 3 \\
N batch size & 32 & 64 & 64 & 64 & 32 & 64 & 32 & 64 & 32 \\
N hidden & 1024 & 1024 & 1024 & 1024 & 1024 & 1024 & 1024 & 1024 & 1024 \\
N  layers & 4 & 6 & 6 & 6 & 4 & 6 & 4 & 6 & 4 \\
N volume & 180000 & 180000 & 180000 & 180000 & 180000 & 180000 & 180000 & 180000 & 180000 \\
E batch size & 8 & 8 & 8 & 32 & 32 & 32 & 8 & 32 & 32 \\
E hidden & 512 & 512 & 512 & 512 & 512 & 512 & 512 & 512 & 512 \\
E layer & 4 & 4 & 4 & 4 & 4 & 4 & 4 & 4 & 4 \\
E volume & 25000 & 25000 & 25000 & 50000 & 50000 & 50000 & 30000 & 50000 & 25000 \\
E oversampling $OS\%$ & 1.0 & 1.0 & 1.0 & 1.0 & 1.0 & 1.0 & 1.0 & 1.0 & 1.0 \\
C batch size & 64 & 8 & 64 & 64 & 8 & 64 & 8 & 64 & 64 \\
C hidden & 1024 & 1024 & 1024 & 1024 & 1024 & 1024 & 1024 & 1024 & 1024 \\
C  layers & 4 & 4 & 4 & 4 & 4 & 4 & 4 & 4 & 4 \\
C volume & 100000 & 30000 & 100000 & 100000 & 30000 & 100000 & 30000 & 100000 & 100000 \\
N oversampling $OS\%$ & 1.0 & 1.0 & 1.0 & 1.0 & 1.0 & 1.0 & 1.0 & 1.0 & 1.0 \\
Loss-alpha & 2 & 2 & 3 & 1 & 1 & 1 & 2 & 2 & 1 \\
Loss-beta & 0.5 & 0.5 & 0.45 & 1 & 1 & 1 & 0.5 & 0.5 & 1 \\
\hline
\end{tabular}
\end{table*}

\section{Dataset and Code}
The data for the 9 reservoirs and 4 rain gauge sensors we used in this study, along with the code for this work have been made available on GitHub at \url{https://github.com/davidanastasiu/NECPlus}.

\section{Discussion}
While trend-based models attempt to capture the non-linear trend and seasonality in time series, the effectiveness of these approaches will be constrained by datasets like reservoir water levels.
Our observations show that the trends are more intricate than in typical time series. Furthermore, as we showed in our work, a single data-driven model struggles to strike a balance between normal and extreme value forecasting due to the rarity and magnitude of severe events. By using selective backpropogation, NEC+ effectively isolates and simplifies the signal trends and values scope. Utilizing RNN and FC layers, NEC+ transforms the problem of predicting continuous time series points into one of first identifying the type of the upcoming point and then forecasting its value using a highly-tuned normal or extreme predictive model. NEC can be updated as a framework by using different neural network implementations for the N, E, and C models separately. Utilizing the same original data source with a constant extreme threshold value $\epsilon$ is a crucial rule to follow. Other statistical methods, such as quantile~\cite{tyralis2021quantile2} , the expectile~\cite{waltrup2015expectile2} information, or other transformation skills~\cite{Yifei2020} can also be used in place of the GMM indicator.

%%%%%%%%%%%%%%%%%%%%%%%%%%%%%%%%%%%%%%%%%%%%%%%%%%%%%%%%%%%%%%%%%%%%%%%%%%%%%%%%

% % {\small
% \bibliography{references}
% % }

\begin{acknowledgements}
Funding for this project was made possible by the Santa Clara Valley Water District.
\end{acknowledgements}

\section*{Bibliography}
\bibliography{arxiv.bib}

\end{document}